\documentclass[nohyperref]{article}

\usepackage{microtype}
\usepackage{graphicx}
\usepackage{booktabs} %
\usepackage{xspace}
\usepackage{xcolor}         %
\usepackage{multirow}
\usepackage{epsfig}
\usepackage{amsmath}
\usepackage{amssymb}

\usepackage{url}
\usepackage{caption}
\usepackage{subcaption}
\usepackage{color}
\usepackage{enumitem}

\usepackage{hyperref}

\usepackage[accepted]{style_files/icml2022}

\usepackage{amsmath,amsfonts,bm}

\def\eqref#1{equation~\ref{#1}}

\def\1{\bm{1}}
\newcommand{\train}{\mathcal{D}}

\DeclareMathAlphabet{\mathsfit}{\encodingdefault}{\sfdefault}{m}{sl}
\SetMathAlphabet{\mathsfit}{bold}{\encodingdefault}{\sfdefault}{bx}{n}

\def\gN{{\mathcal{N}}}

\newcommand{\E}{\mathbb{E}}
\newcommand{\Ls}{\mathcal{L}}
\newcommand{\R}{\mathbb{R}}

\usepackage{style_files/misc_symbols}

\newcommand{\rsim}{$S_r$\xspace}
\DeclareMathOperator{\orsim}{S_r}
\newcommand{\isim}{$S_i$\xspace}
\DeclareMathOperator{\oisim}{S_i}
\newcommand{\stir}{\text{STIR}\xspace}
\DeclareMathOperator{\ostir}{\text{STIR}}
\newcommand{\cka}{\text{CKA}\xspace}
\newcommand{\cca}{\text{CCA}\xspace}
\newcommand{\svcca}{\text{SVCCA}\xspace}
\newcommand{\pwcca}{\text{PWCCA}\xspace}
\newcommand{\iri}{\text{IRI}\xspace}
\newcommand{\iris}{\text{IRIs}\xspace}

\DeclareMathOperator{\merm}{m^{\text{(Vanilla)}}}
\DeclareMathOperator{\madv}{m^{\text{(AT)}}}
\newcommand{\mone}{$m_1$}
\newcommand{\mtwo}{$m_2$}
\newcommand{\rsm}{\text{RSM}\xspace}
\newcommand{\rsms}{\text{RSMs}\xspace}

\icmltitlerunning{Measuring Representational Robustness}

\begin{document}

\twocolumn[
\icmltitle{Measuring Representational Robustness of Neural Networks \\Through Shared Invariances}

\begin{icmlauthorlist}
\icmlauthor{Vedant Nanda}{umd,mpi}
\icmlauthor{Till Speicher}{mpi}
\icmlauthor{Camila Kolling}{pucrs}
\icmlauthor{John P. Dickerson}{umd}
\icmlauthor{Krishna P. Gummadi}{mpi}
\icmlauthor{Adrian Weller}{cam,tur}
\end{icmlauthorlist}

\icmlaffiliation{umd}{University of Maryland, College Park}
\icmlaffiliation{mpi}{Max Planck Institute for Software Systems}
\icmlaffiliation{pucrs}{PUCRS}
\icmlaffiliation{tur}{The Alan Turing Institute}
\icmlaffiliation{cam}{University of Cambridge}

\icmlcorrespondingauthor{Vedant Nanda}{vedant@cs.umd.edu}

\icmlkeywords{Machine Learning, ICML}

\vskip 0.3in
]

\printAffiliationsAndNotice{}  %

\begin{abstract}
A major challenge in studying robustness in deep learning is defining the set of ``meaningless'' perturbations to which a given Neural Network (NN) should be invariant. Most work on robustness implicitly uses a human as the reference model to define such perturbations. Our work offers a new view on robustness by using another reference NN to define the set of perturbations a given NN should be invariant to, thus generalizing the reliance on a reference ``human NN'' to any NN. This makes measuring robustness equivalent to measuring the extent to which two NNs share invariances, for which we propose a measure called \stir. \stir re-purposes existing representation similarity measures to make them suitable for measuring shared invariances. 
Using our measure, we are able to gain insights into how shared invariances vary with changes in weight initialization, architecture, loss functions, and training dataset. Our implementation is available at: \url{https://github.com/nvedant07/STIR}.
\vspace{-4mm}
\end{abstract}

\section{Introduction}\label{sec:intro}

As deep neural networks are increasingly deployed in real-world scenarios, robustness of their automatically learned feature representations has emerged as a key desideratum. Prior works have defined many measures of robustness corresponding to different types of synthetically-generated or naturally-occurring perturbations that can be applied to the inputs and  their distributions (e.g., adversarial inputs \cite{biggio2013evasion,szegedy2013intriguing,papernot2016limitations} or distributional shifts \cite{geirhos2018generalisation,hendrycks2018benchmarking,engstrom2019exploring,alhussein2015classifiers,taori2020measuring}).
The common property underlying the different robustness measures is that they all attempt to capture \emph{the extent to which the learned representations remain invariant (i.e., unchanged) under some defined set of perturbations}.

In this paper, we propose a new way to study, understand, and characterize robustness of neural networks. 
Our key insight is that the set of input perturbations against which a neural network's robustness is measured, can itself be defined by another \emph{reference} neural network. 
Specifically, given a reference neural network, we first obtain a set of input perturbations that are imperceptible to the reference network (i.e., find inputs with invariant reference representations), and then check the extent to which representations of other neural networks are invariant to these perturbations.
Our proposal allows us to measure relative invariance of two neural network representations and estimate the degree to which the two neural networks share representational invariance.
Intuitively, our proposal generalizes the often unstated, but implicit assumption behind all interesting sets of perturbations used in robustness studies today: they are perturbations that are imperceptible to a particular reference neural network, the human brain.

Comparing representational invariance of two neural networks is an important aspect of determining their representational (or perceptual) alignment.
Assessing representational alignment is crucial for a future society with interacting agents controlled by neural networks (e.g., cars driven by different deep learning systems).
Additionally, the ability to measure relative invariance, and therefore robustness, of deep neural network representations can offer insights into interesting questions such as: when updating a model, to what extent are invariances preserved (which may be crucial to regulators for safety assurance)? How does representational invariance vary with the choice of network architectures, loss functions, random weight initialization, and datasets used in the training process?

Our work is inspired by and builds upon previous studies investigating similarity between deep neural network representations \cite{raghu2017svcca,morcos2018insights,kornblith2019similarity,nguyen2020wide}.
However, we find that existing representational similarity literature focuses \emph{narrowly} on comparing two representations of data samples drawn from a specific input distribution, ignoring representations of data samples outside of the distribution or changes in representation caused by input perturbations for which one of the two representations remains invariant.
As such, existing measures of similarity between neural network representations offer \emph{no} insight into their robustness and consequently, their alignment.
Nevertheless, we retain the compelling (axiomatic) properties of existing similarity measures, by re-purposing them to measure relative invariance.
Specifically, for input perturbations imperceptible to the reference neural network, we quantify the invariance in representations of the other neural network using a popular similarity index called Centered Kernel Alignment (CKA) \cite{kornblith2019similarity}.

To summarize, our key contributions are as follows:

\begin{itemize}[leftmargin=*]
    \item We propose a measure of shared invariances between two representations that is based on the models which generated them. We show that our measure faithfully captures shared invariances between two models where existing measures of representation similarity (such as CKA) do not work adequately.
    \item Our proposal repurposes existing representation similarity measures to measure shared invariance, thus preserving all the (desirable) axiomatic properties of these measures. 
    \item Using our measure we are able to derive novel insights about the impact of weight initialization, architecture, loss and training dataset on the shared invariances between networks. Our initial results show that our measure is a promising evaluation tool to better understand deep learning.
    \item We find that typically the shared invariance between models reduces for later layers, however when trained using adversarial training, the same models end up with higher shared invariances even in later layers. We also see that models with residual connections tend to have high shared invariances among them than between other non-residual models.
\end{itemize} %
\section{Measuring Representational Robustness}
\label{sec:robustness}

Robustness is defined as ``perform(ing) without failure under a wide range of conditions'' \cite{mw22robust}. Applying the definition in the context of learning models, we can disentangle two distinct requirements for a model to be considered robust: i) the model must produce correct outputs, \ie~have high accuracy, and ii) these outputs must be produced consistently for a diverse set of inputs, \ie~the outputs of the model must be \textit{invariant} to irrelevant perturbations (changes) in the input. Extensive literature on robust learning~\cite{szegedy2013intriguing,papernot2016limitations,goodfellow2015explaining} suggests that it is quite hard to train models that achieve high accuracy on standard benchmarking datasets and high invariance to irrelevant (adversarially-generated or naturally-occurring) perturbations simultaneously~\cite{tsipras2018robustness,madry2019deep,zhang2019theoretically}. Reconciling correctness and invariance requirements remains a topic of active research.

Here, we investigate the robustness of neural network representations that are learned in the process of generating outputs.
Specifically, we attempt to quantify the invariance of learned neural network representations to \textit{irrelevant} perturbations in the inputs.
Intuitively, a high representational invariance is a necessary (though not sufficient) condition for a neural network model to be robust.

\subsection{A Relative Invariance Framework}

In order to quantify the invariance of a neural network's representations to irrelevant perturbations to inputs, we need to first define the set of such perturbations. Most of the existing works on robustness use humans as a reference model to define these perturbations. Thus all works on robustness are inherently (and often implicitly) \textit{relative} to a human. We generalize the reliance on a humans as the reference model by assuming the reference model to be another neural network. We can then define the set of irrelevant input perturbations as those changes that do not cause any change in the reference model's representation (i.e., perception) of the inputs. Finally, we quantify how invariant a given neural network's representations are to these irrelevant input perturbations.

Our framework effectively quantifies invariance of one neural network's representation relative to another reference neural network. Our use of a reference neural network model is inspired by how human perception is used as a reference to determine which adversarial perturbations~\cite{szegedy2013intriguing} or image corruptions \cite{geirhos2018generalisation,hendrycks2018benchmarking} or image transformations \cite{engstrom2019exploring,alhussein2015classifiers} do not alter the perception of inputs and are, hence, considered irrelevant perturbations.

\subsection{Problem Setting}\label{sec:problem}

Given a reference neural network $m_1: \R^{m} \mapsto \R^{d1}$ and $n$ samples ($X \in \R^{n \times m}$) from a given data distribution ($\train$), our goal is to define a measure of how invariant a given target network $m_2: \R^{m} \mapsto \R^{d2}$ is to perturbations of samples $X$ that are imperceptible by $m_1$, \ie~do not change their representations according to $m_1$.

\paragraph{Invariant input perturbations for a reference model $(X')$}

In our framework, the invariances that are desirable are determined with respect to a reference model, $m_1$. To this end, we introduce the notion of \textit{Identically Represented Inputs (\iris)}. For any given input data point $x$ and a given reference model $m_1$, \iris is the set of all data points that are mapped to the same representation as $x$ by $m_1$. Formally,

\begin{equation}
    \iri_{\text{strict}}(x;m_1) = \{x' \,\, | \,\, m_1(x') = m_1(x)\}
\end{equation}

In other words, $m_1$ is \textit{invariant} to and cannot perceive any difference between $x$ and any $x' \in \iri(x;m_1)$. In practice, exact equality is hard to achieve, and thus we relax this formulation so that $x$ and $x'$ are \textit{almost} indistinguishable for $m_1$, \ie, for a small enough $\delta$,
\begin{equation}
    \iri_{\text{relax}}(x;m_1) = \{x' \,\, | \,\, \frac{||m_1(x') - m_1(x)||_2}{|| m_1(x) ||_2} \leq \delta\}
\end{equation}

Going forward we use the relaxed formulation of \iris and thus omit the subscript. For each of the $n$ input points $x \in X$, if we pick a $x'$ from $\iri(x;m_1)$, we get a corresponding batch of $n$ samples $X'$ such that $m_1(X') \approx m_1(X)$.

\paragraph{Measuring invariance of the target model on $(X, X')$}
Assuming we obtain $X$ and $X'$, our key idea is to capture the extent to which $m_1$ and $m_2$ share invarances, by measuring the degree to which representations assigned by $m_2$ to $X$ and $X'$ are \textit{similar}. Specifically, we want to quantify the degree of similarity between two sets of $n$ data representations $Y = m_2(X) \in \R^{n \times d2}$ and $Z = m_2(X') \in \R^{n \times d2}$. To this end, we can make use of any of the existing representation similarity measures (\rsim), such as CKA~\cite{kornblith2019similarity}, and variants of CCA~\cite{morcos2018insights,raghu2017svcca} as they're designed specifically to measure similarity between two sets of representations. 
This yields a \textit{shared invariance measure, \isim}, which can now be formally defined as:

\begin{equation}\label{eq:isim_def}
    \begin{split}
    \oisim(m_2 | m_1, X, X', \orsim) = \orsim(m_2(X), m_2(X'))
    \end{split}
\end{equation}

Note that while the traditional use of \rsim, \ie, $\orsim(m_1(X), m_2(X))$ does not measure shared invariance between $m_1$ and $m_2$ (we give concrete arguments for the same in Section~\ref{sec:sim_problems_logical}), our proposed measure (Eq~\ref{eq:isim_def}) shows how existing \rsim measures can be repurposed to measure shared invariance.

It's important to note that our measure is a directional one since \iris are defined relative to the reference model $m_1$. Other than the reference and target models $m_1$ and $m_2$, our measure takes given input points $X$ and a representation similarity measure (\rsim) as inputs. We discuss the concrete instantiations of $\oisim$ used in this work in Section \ref{sec:stir_def}. First, however, we describe how to construct $X'$ given $X$.

\subsection{Generating \iris}
\label{sec:generating_iri}

Operationalizing Eq~\ref{eq:isim_def}, requires the answer to a key question: \textbf{how to sample $x'$ from an infinitely large set $\iri(x;m_1)$?} We argue that there are two key ways to do this: \textit{arbitrarily} or \textit{adversarially}. We can randomly choose a sample from $\iri(x;m_1)$, in which case we get \textit{arbitrary} \iris, or we can pick $x'$ adversarially with respect to $m_2$, such that the representations $m_2(x')$ and $m_2(x)$ are farthest apart, in which case we get \textit{adversarial} \iris. 
We also show in Table~\ref{tab:walkthorugh_stir} that takeaways about shared invariance can vary greatly depending on the choice of arbitrary or adversarial \iris.

We leverage the key insight that $m_1$ can map multiple different inputs to the same representation (since $m_1$ is a highly non-linear deep neural network) which can be found using representation inversion~\cite{mahendran2014understanding}. For a given set of inputs typically drawn from the training distribution $X = [x_1 ... x_n]$ and a reference model $m_1$, we generate $X' = [x'_1 ... x'_n]$ that are all mapped to similar representations as $X$ by $m_1$, \ie~$m_1(X) \approx m_1(X')$. Note that $X$ and $X'$, are, by construction \iris. This is achieved by performing the following optimization for every $x \in X$:
\begin{equation}\label{eq:rep_inv}
    \text{argmin}_{x'}  \Ls(x') ,
\end{equation}
Which can be approximated using gradient descent by repeatedly performing the following update (where $\alpha$ is the step size):
\begin{equation}\label{eq:rep_inv_gd}
    x' = x' - \alpha * \nabla_{x'} \Ls .
\end{equation}
A key consideration in solving this optimization is that we must start with some initial value of $x'$, \ie~we must choose a seed $x'$ from which to start the gradient descent. Different seeds can lead to different solutions of $x'$. We find that in practice randomly picked seeds give fairly stable estimates\footnote{Since we deal with images we sample each pixel value as a random integer from $0-255$ with uniform probabilities.}. 

\textbf{Arbitrary \iri} In order to simulate an arbitrary sample from $\iri(x;m_1)$, we use the following $\Ls$,

\begin{equation*}
    \Ls_{\text{arbitrary}}(x') = || m_1 (x') - m_1 (x) ||_2 .
\end{equation*}

Since this scheme only optimizes for similar representation of $x$ and $x'$ on $m_1$ ($\forall x \in X$) and starts from a random initial value of $x'$, it simulates a random sample from the (possibly infinitely) large set $\iri(x;m_1)$.

\textbf{Adversarial \iris} We can alter the way we find $x'$ ($\forall x \in X$) such that the resulting ($X, X'$) are still, by definition \iris but are optimized to generate very distinct outputs on $m_2$, the model for which we're measuring shared invariance. This can be achieved by using exactly the same procedure as in Eqs~\ref{eq:rep_inv} and \ref{eq:rep_inv_gd}, except we now change $\Ls$ to:
\begin{equation*}
    \begin{split}
        \Ls_{adv} = || m_1(x') - m_1(x) ||_2 - \\|| m_2(x') - m_2(x) ||_2 .   
    \end{split}
\end{equation*}
Solving for $x'$ using $\Ls_{adv}$ ensures that the inputs are still similarly represented as $x$ on $m_1$ ($\forall x \in X$) and thus (X, X') are \iris. However this ensures that any measure of \isim on such \iris will be a worst case estimate. Such \iris have been referred to as \textit{controversial stimuli}~\cite{golan2020controversial} in existing literature.

\section{Measuring Shared Invariances}
\label{sec:stir}

\def\arraystretch{1.25}
\begin{table*}[!t]
\begin{center}
\begin{small}
\begin{sc}
\begin{tabular}{c|cc|cc|cc}
\hline

& \multicolumn{2}{c|}{\begin{tabular}[c]{@{}c@{}}ResNet18 Vanilla (\textbf{$m_1$}),\\ResNet18 Vanilla (\textbf{$m_2$}) \end{tabular}} & \multicolumn{2}{c|}{\begin{tabular}[c]{@{}c@{}}ResNet18 AT (\textbf{$m_1$}),\\ResNet18 AT (\textbf{$m_2$}) \end{tabular}} & \multicolumn{2}{c}{\begin{tabular}[c]{@{}c@{}}ResNet18 AT (\textbf{$m_1$}),\\ResNet18 Vanilla (\textbf{$m_2$}) \end{tabular}} \\
 & & & & & \\
\hline

 & \multirow{2}{*}{\begin{tabular}[c]{@{}c@{}}\bf{$m_1 | m_2$}\end{tabular}} & \multirow{2}{*}{\begin{tabular}[c]{@{}c@{}}\bf{$m_2 | m_1$}\end{tabular}} &  \multirow{2}{*}{\begin{tabular}[c]{@{}c@{}}\bf{$m_1 | m_2$}\end{tabular}} & \multirow{2}{*}{\begin{tabular}[c]{@{}c@{}}\bf{$m_2 | m_1$}\end{tabular}} &  \multirow{2}{*}{\begin{tabular}[c]{@{}c@{}}\bf{$m_1 | m_2$}\end{tabular}} & \multirow{2}{*}{\begin{tabular}[c]{@{}c@{}}\bf{$m_2 | m_1$}\end{tabular}} \\ 
 & & & & & \\ \hline

\stir & $0.605_{\pm 0.013}$ & $0.562_{\pm 0.023}$ & $0.934_{\pm 0.003}$ & $0.939_{\pm 0.002}$ & $0.405_{\pm 0.020}$ & $0.509_{\pm 0.011}$ \\ 
$\ostir_{adv}$ & $0.085_{\pm 0.004}$ & $0.064_{\pm 0.004}$ & $0.096_{\pm 0.007}$ & $0.078_{\pm 0.005}$ & $0.054_{\pm 0.004}$ & $0.070_{\pm 0.004}$ \\ 
\cka & \multicolumn{2}{c|}{$0.967_{\pm 0.000}$} & \multicolumn{2}{c|}{$0.937_{\pm 0.000}$} & \multicolumn{2}{c}{$0.536_{\pm 0.000}$} \\
Acc($m_1(X')$, $m_2(X')$) & $0.521_{\pm 0.061}$ & $0.347_{\pm 0.029}$ & $0.891_{\pm 0.007}$ & $0.901_{\pm 0.002}$ & $0.140_{\pm 0.028}$ & $0.555_{\pm 0.012}$ \\

\end{tabular}
\end{sc}
\end{small}
\end{center}

\caption{
\textbf{[\stir faithfully estimates shared invariance]} Here the two ResNet18s in each column are trained on CIFAR10 with different random initializations, holding every other hyperparameter constant. 1.) For two such models trained using the vanilla crossentropy loss (left), interestingly, we find that \stir highlights a lack of shared invariance, whereas \cka overestimates this value; 2.) when both models are trained using adversarial training~\cite{madry2019deep} (middle) \stir faithfully estimates high shared invariance; 3.) Finally \stir is able show how having a directional measure can bring out the differences when comparing a model trained with vanilla loss and adv training (right), whereas \cka being unidirectional cannot derive these insights. All numbers are computed over 1000 random samples from CIFAR10 training set and averaged over 5 runs.}
\label{tab:walkthorugh_stir}
\vspace{-4mm}
\end{table*}

\subsection{\stir, an instantiation of $\oisim$}\label{sec:stir_def}

Using Eqs~\ref{eq:rep_inv}\&\ref{eq:rep_inv_gd} we can generate $k$ different $(X, X')$, by repeatedly sampling $X$ $k$ times from a given distribution (typically the training distribution of $m_1$, \eg~the train or test set). 
Now, we define $\ostir(m_2 | m_1, X, \orsim)$ 
as follows:
\begin{equation}\label{eq:stir}
    \ostir(m_2 | m_1, X, \orsim) = \frac{1}{k}\sum_{X'}\orsim(m_2(X), m_2(X')) .
\end{equation}
Here, we find $X'$ using representation inversion as described in Section \ref{sec:generating_iri}.
We call this measure \textbf{S}imilarity \textbf{T}hrough \textbf{I}nverted \textbf{R}epresentations --- \stir. 

When all $X'$ are chosen in an adversarial manner (\ie using $\Ls_{adv}$ as described in Section~\ref{sec:generating_iri}), we can estimate the worst case \stir as:

\begin{equation}\label{eq:stir_adv}
    \ostir_{adv}(m_2 | m_1, X, \orsim) = \frac{1}{k}\sum_{X^{'}_{adv}}\orsim(m_2(X), m_2(X')) .
\end{equation}

Both Equations~\ref{eq:stir}\&\ref{eq:stir_adv} are parametrized by \rsim. For our purpose we use linear Centered Kernel Alignment (\cka)~\cite{kornblith2019similarity} as the similarity measure, \ie \rsim $=$ Linear \cka. \cka has been adopted by the community as the standard measure for representation similarity and has been used by many subsequent works to derive important insights about deep learning~\cite{nguyen2020wide,raghu2021vision,neyshabur2020being}. \cka also has certain desirable axiomatic properties such as invariance to isotropic scaling and orthogonal transformations, which other methods such as \svcca~\cite{raghu2017svcca} and \pwcca~\cite{morcos2018insights} do not possess.
Importantly, \cka is \textit{not} invariant to every invertible linear transformation, which is not true of \svcca or \pwcca.
We refer to the paper~\cite{kornblith2019similarity} for an extensive discussion on why these are desirable properties for any similarity measure.
While CKA, as proposed, can work with any kernel function, results show that Linear CKA works just as well as RBF CKA, so for simplicity we use Linear CKA for all our experiments.
Definitions of the similarity metrics listed above can be found in Appendix \ref{sec:appendix_similarity_measures}.

\subsection{Why Existing Representation Similarity (\rsim) Measures Cannot Measure Shared Invariance}
\label{sec:sim_problems_logical}

While at first glance a representation similarity measure (\rsm) such as \cka \cite{kornblith2019similarity} or one of the others mentioned in section \ref{sec:stir_def} might look appealing to measure shared invariance, these measures are in fact not suitable for this task, for several reasons.

First, \textbf{current \rsms are not designed for capturing invariance.}
Their goal is to measure the degree of correlation between sets of points generated by two different models, that are transformed to be as aligned as possible under certain constraints.
Measuring invariance, however, requires capturing the degree of difference between points generated \textit{by the same model}, which should be the same.

Second, \textbf{\rsms don't interact with the model.}
All existing \rsms are evaluated in a two-step process, by first obtaining collections of representations $Y = m_1(X), Z = m_2(X)$ from two models and then computing similarity based on those representations as $\orsim(Y, Z)$, without using the models further.
From a causal perspective, an invariance is an intervention on a model's input that does not lead to a change in the model's output.
However, a central result in the causality literature is that the effect of interventions cannot be determined from observational data alone \cite{pearl09causality}.
Therefore, any metric that aims to make meaningful statements about model invariance needs to interact with the model to make interventions.
\stir does this via the process of representation-inversion, however, existing \rsms are purely observational and therefore cannot properly determine invariances.

Third, \textbf{sharing invariance between two models is directional.}
If model $m_1$ is constant, it will share all of model $m_2$'s invariances, but not vice versa if $m_2$ is not constant itself.
Existing \rsms are not directional and therefore cannot express these relationships.

\subsection{\stir Faithfully Measures Shared Invariance}
\label{sec:sim_problems_empirical}

Training two models ($m_1$ and $m_2$) with different random initializations (holding all other things like architecture, loss and other hyperparameters constant), leads to very ``similar'' representations on the penultimate layer, as measured by instantiations of \rsim such as \cka~\cite{kornblith2019similarity}. We consider two variants of this experiment, where we train two ResNet18 models (on CIFAR10) from different random initializations (keeping everything else same) with 1.) the standard crossentropy loss ($\merm_1$, $\merm_2$), and 2.) with adversarial training ($\madv_1$, $\madv_2$)~\footnote{trained using $\ell_2$ threat model with $\epsilon = 1.0$, see~\cite{madry2019deep} for more details}. Throughout the paper \stir is measured over CIFAR10 training samples with \rsim = Linear CKA. Thus, to simplify the notation, we use $\ostir(m_1 | m_2)$ to mean $\ostir(m_1 | m_2, X \sim \text{CIFAR10}, \text{Linear} \cka)$.

\textbf{\stir provides insights into shared invariance where \cka fails.}
Both ($\merm_1$, $\merm_2$) and ($\madv_1$, $\madv_2$) achieve a high similarity score (as measured by \cka, on the penultimate layer of both models).
However, we find that such a similarity measurement would be an overestimation of shared invariances between $\merm_1$ and $\merm_2$ which are much lower when measured as $\ostir(\merm_1 | \merm_2)$ and $\ostir(\merm_2 | \merm_1)$, as shown in  Table~\ref{tab:walkthorugh_stir}.
Two models trained using adversarial training ($\madv_1$, $\madv_2$) should intuitively have more shared invariances since these models were explicitly trained to be invariant to $\ell_2$ perturbations. Indeed, we see that these two models have much higher values of $\ostir(\madv_1 | \madv_2)$ and $\ostir(\madv_2 | \madv_1)$.

\textbf{Sanity Checks for \stir}
For high values of \stir, intuitively we would expect the representations of the model we're evaluating ($m_2$) to have similar representations on \iris, \ie~$m_2(X) \approx m_2(X')$. Since \iris, by construction, have similar representation on the reference model ($m_1$), we expect the predictions of $m_2$ on $X'$ (pred($m_2(X')$) to agree with predictions of $m_1$ on $X'$ (pred($m_1(X')$). Similarly, for low \stir values, we'd expect less agreement between pred($m_1(X')$) and pred($m_2(X')$). We see that this relationship holds as lower \stir values for $\merm_1$ and $\merm_2$ also correspond to less agreement in their predictions on $X'$ and higher \stir values for $\madv_1$ and $\madv_2$ correspond to higher agreement in their predictions (as shown in the right of each row of Table~\ref{tab:walkthorugh_stir}).
To further corroborate that the estimate of shared invariance given by \stir is justified in being lower for ($\merm_1$, $\merm_2$) than ($\madv_1$, $\madv_2$), we generate \textit{controversial stimuli}~\cite{golan2020controversial} for both pairs of models. Details of how to generate these can be found in the Appendix~\ref{sec:appendix_faithful_stir}.
We find that indeed it's significantly easier to generate controversial stimuli for ($\merm_1$, $\merm_2$) than for ($\madv_1$, $\madv_2$) (see Appendix~\ref{sec:appendix_faithful_stir} for the results). 

\textbf{$\ostir_{adv}$ shows that in the worst case, there are almost no shared invariances.} When measuring shared invariance in the worst case (using $\ostir_{adv}$, Eq~\ref{eq:stir_adv}), we see that even in the case of $\madv_1$ and $\madv_2$ the shared invariance drops close to $0$. This is shown in the second row of Table~\ref{tab:walkthorugh_stir}. Thus, for the rest of the paper we focus on STIR measured using arbitrary \iris, since $\ostir_{adv}$ gives a very pessimistic estimate of shared invariance and thus is not useful in comparing models.

\textbf{\stir brings out nuance through directionality.}
When comparing across training types, \eg~($\madv_1$, $\merm_2$), we find that directionality of \stir is able to show that invariances of $\merm_2$ are not well captured by $\madv_1$ (indicated by the low value of $\ostir(\merm_1 | \madv_2)$). However, $\ostir(\madv_2 | \merm_1)$ is much higher. 
We posit that models trained using AT have a ``superior'' set of invariances and do not posses ``bad'' invariances that models trained using the vanila loss exhibit. Thus, when evaluating \iris from a Vanilla model on a model trained using AT -- one can expect lower shared invariance than in the other direction. \stir is able to capture these nuances in comparison between models because of its directionality. Measures of representation similarity (\rsim) like \cka do not offer any such insights, since they're not directional.

\section{Using \stir to Analyze Model Updates}\label{sec:stir_uses}

One motivation for a shared invariance measure like \stir, as mentioned in Section~\ref{sec:intro}, is to monitor how different models ``align'' with each other. This can then be used to analyze if updates to a model leads to preservation of invariances learned before the update. We first show that \stir can capture relative differences between model invariances and then use \stir to analyze a simulated model update scenario where we incrementally add more training data.

\subsection{\stir Captures Relative Robustness}
We demonstrate that \stir can capture differences between models of varying degrees of robustness. We know that increasing adversarial robustness increases invariance to $\ell_p$ ball perturbations. Thus, by construction, if a model $m_1$ has higher adversarial robustness than another model $m_2$, then invariances of $m_1$ should be ``superior'' to those of $m_2$ and hence we should see $\stir(m_2|m_1) > \stir(m_1|m_2)$. To empirically test this, we construct three models with varying degrees of adversarial robustness: a model trained using the vanilla crossentropy loss ($0\%$), a model trained using adversarial training (AT) with the usual 10 iterations used to solve the inner maximization of AT~\cite{madry2019deep} ($51.5\%$), and a model trained using AT but with only 1 iteration in the inner maximization loop, which gives adversarial robustness somewhere in the middle ($34.6\%$). Table~\ref{tab:stir_robustness} shows comparison between these three models and confirms that $\stir(m_2|m_1) > \stir(m_1|m_2)$ if adversarial robustness $m_1 > m_2$ (measured here by robust accuracy).

\begin{table}[!t]
\begin{center}
\begin{small}
\begin{sc}
\begin{tabular}{c|c|c|c}
\hline

 & \multirow{2}{*}{\begin{tabular}[c]{@{}c@{}}AT \\it=10\end{tabular}} & \multirow{2}{*}{\begin{tabular}[c]{@{}c@{}}AT \\it=1\end{tabular}} & \multirow{2}{*}{\begin{tabular}[c]{@{}c@{}}Vanilla \end{tabular}} \\
 & & \\\hline

 \multirow{3}{*}{\begin{tabular}[c]{@{}c@{}}\textbf{AT it=10} \\Rob: $51.5_{\pm 0.6}$ \\Clean: $80.4_{\pm 0.4}$ \end{tabular}} & \multirow{3}{*}{---} & \multirow{3}{*}{\begin{tabular}[c]{@{}c@{}}$0.93 \pm_{0.01}$\\$(_{0.89 \pm 0.02})$\end{tabular}} & \multirow{3}{*}{\begin{tabular}[c]{@{}c@{}}$0.51 \pm_{0.01}$\\$(_{0.56 \pm 0.01})$\end{tabular}}\\
 & & & \\
 & & & \\\hline
 
 \multirow{3}{*}{\begin{tabular}[c]{@{}c@{}}\textbf{AT it=1}\\Rob: $34.6_{\pm 0.6}$ \\Clean: $87.2_{\pm 2.1}$ \end{tabular}} & \multirow{3}{*}{\begin{tabular}[c]{@{}c@{}}$0.86 \pm_{0.02}$ \\ $(_{0.89 \pm 0.02})$\end{tabular}} & \multirow{3}{*}{---} & \multirow{3}{*}{\begin{tabular}[c]{@{}c@{}}$0.52 \pm_{0.01}$\\$(_{0.64 \pm 0.00})$\end{tabular}} \\
 & & & \\
 & & & \\\hline
 
 \multirow{3}{*}{\begin{tabular}[c]{@{}c@{}}\textbf{Vanilla}\\Rob: $0.0_{\pm 0.0}$ \\ Clean: $95.0_{\pm 0.1}$ \end{tabular}} & \multirow{3}{*}{\begin{tabular}[c]{@{}c@{}}$0.41 \pm_{0.02}$\\$(_{0.56 \pm 0.01})$\end{tabular}} & \multirow{3}{*}{\begin{tabular}[c]{@{}c@{}}$0.34 \pm_{0.05}$\\$(_{0.64 \pm 0.00})$\end{tabular}} & \multirow{3}{*}{---} \\
 & & & \\
 & & & \\\hline

\end{tabular}
\end{sc}
\end{small}
\end{center}
\vspace{-4mm}
\caption{
\texttt{ITERS} is the number of iterations in the inner loop of Adversarial Training (AT). 
Each cell shows $\ostir(m_j|m_i)$ where $m_j$ is the model on the column and $m_i$ is the model on the row. CKA numbers are shown in (). The three models have robust accuracies (measured under $\ell_2, \epsilon = 1$) such that $AT, IT = 10 > AT, IT = 1 > \text{Vanilla}$. \label{tab:stir_robustness}}
\vspace{-7mm}
\end{table}

\subsection{Updating Models With More Data}
\begin{figure}[b!]
    \vspace{-9mm}
    \subfloat[Standard Training]{\label{fig:rob}
        \includegraphics[width=0.7\linewidth]{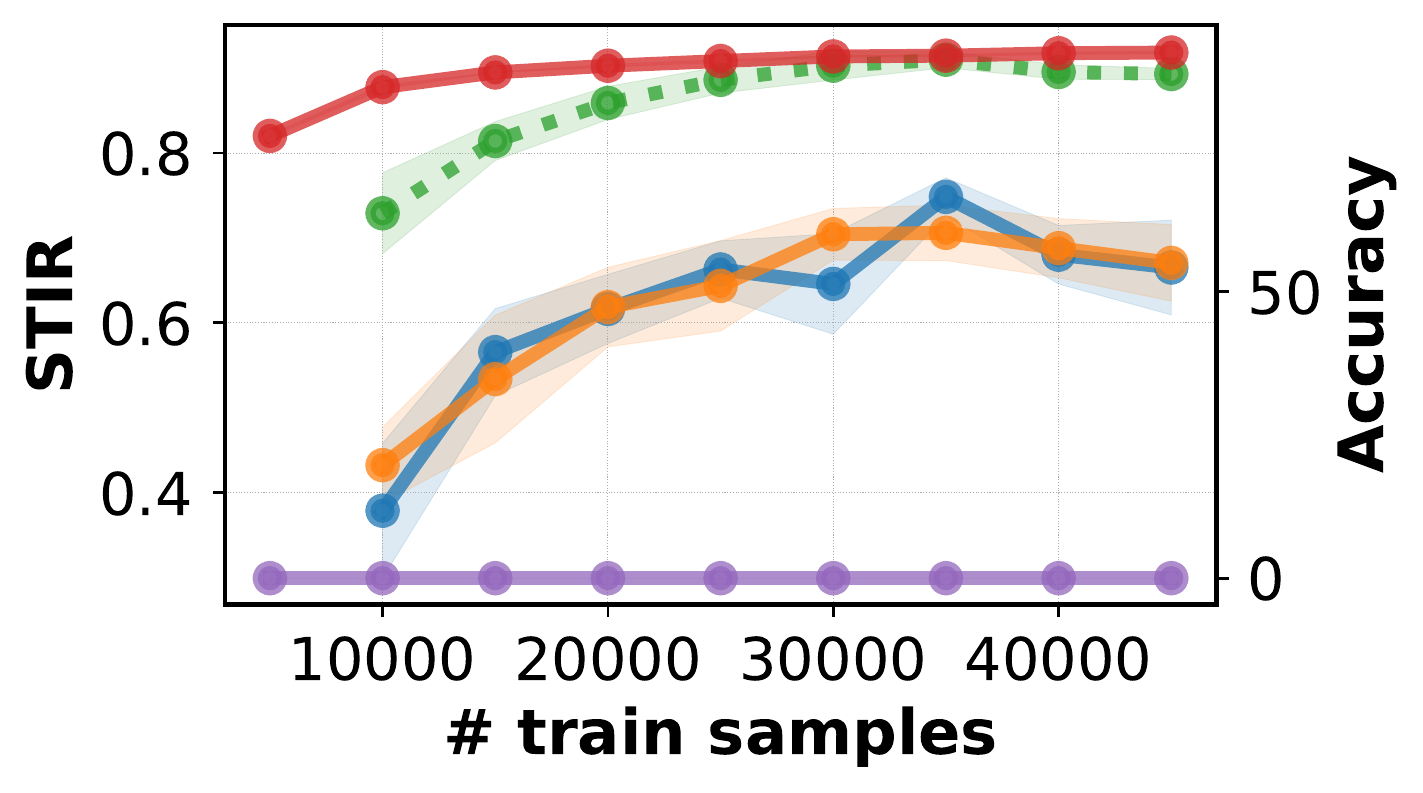}}
    
    \subfloat[Adv Training]{\label{fig:nonrob}
        \includegraphics[width=0.7\linewidth]{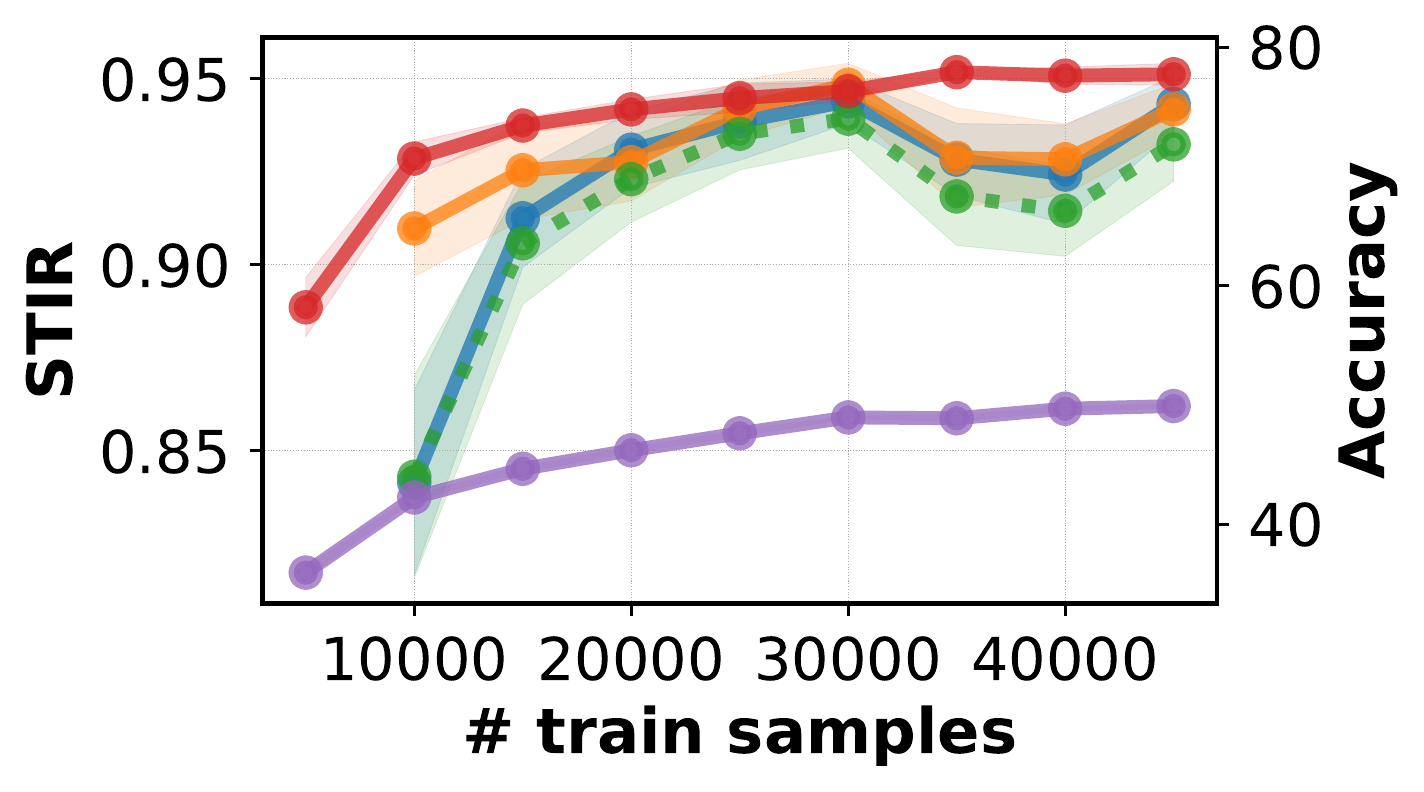}}
    \subfloat{\raisebox{0.04\textheight}{
        \includegraphics[width=0.25\linewidth]{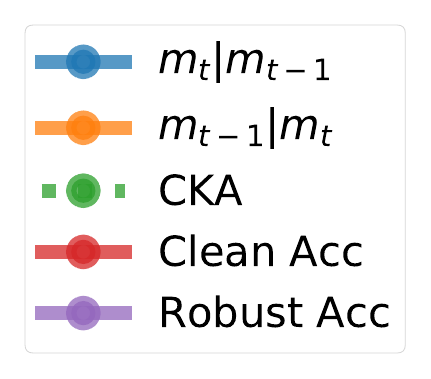}}}
    
\caption{Adding more training data for (x-axis) leads to a monotonic increase in STIR between model at a given timestep ($m_t$) and the previous timestep ($m_{t-1}$) (in both directions).\label{fig:updating_model}}
\end{figure}

Models deployed in the real world are continuously updated with more data. In such cases, it may be crucial to understand how a model update at a given timestep shares invariance with the model at the previous timestep. To simulate such a scenario, we train a ResNet18 on CIFAR10 where at each timestep, we add $5k$ training samples, \ie, at timesteps $t = 0, 1, ..., T$, we train the model on $5k, 10k, ..., 45k$ samples (we keep $5k$ samples for a holdout validation set). At each timestep we train for 100 epochs. Figure~\ref{fig:updating_model} shows how $\ostir(m_t | m_{t-1})$ (and $\ostir(m_{t-1} | m_t)$) changes as we progressively add more data. Here $m_t$ is the model at a given timestep and $m_{t-1}$ is the model on the previous timestep. For both AT and Vanilla loss we see $\ostir(m_t | m_{t-1})$ and $\ostir(m_{t-1} | m_t)$ increase as we add more data. We also see that after a certain amount of data the \stir scores plateau -- thus indicating that adding more data has diminishing returns for shared invariance.

\begin{figure*}[h!]
    \centering
    
    \subfloat[2 ResNet18 with different init, Vanilla]{\label{fig:rand_init_r18nonrob}
        \includegraphics[width=0.3\linewidth]{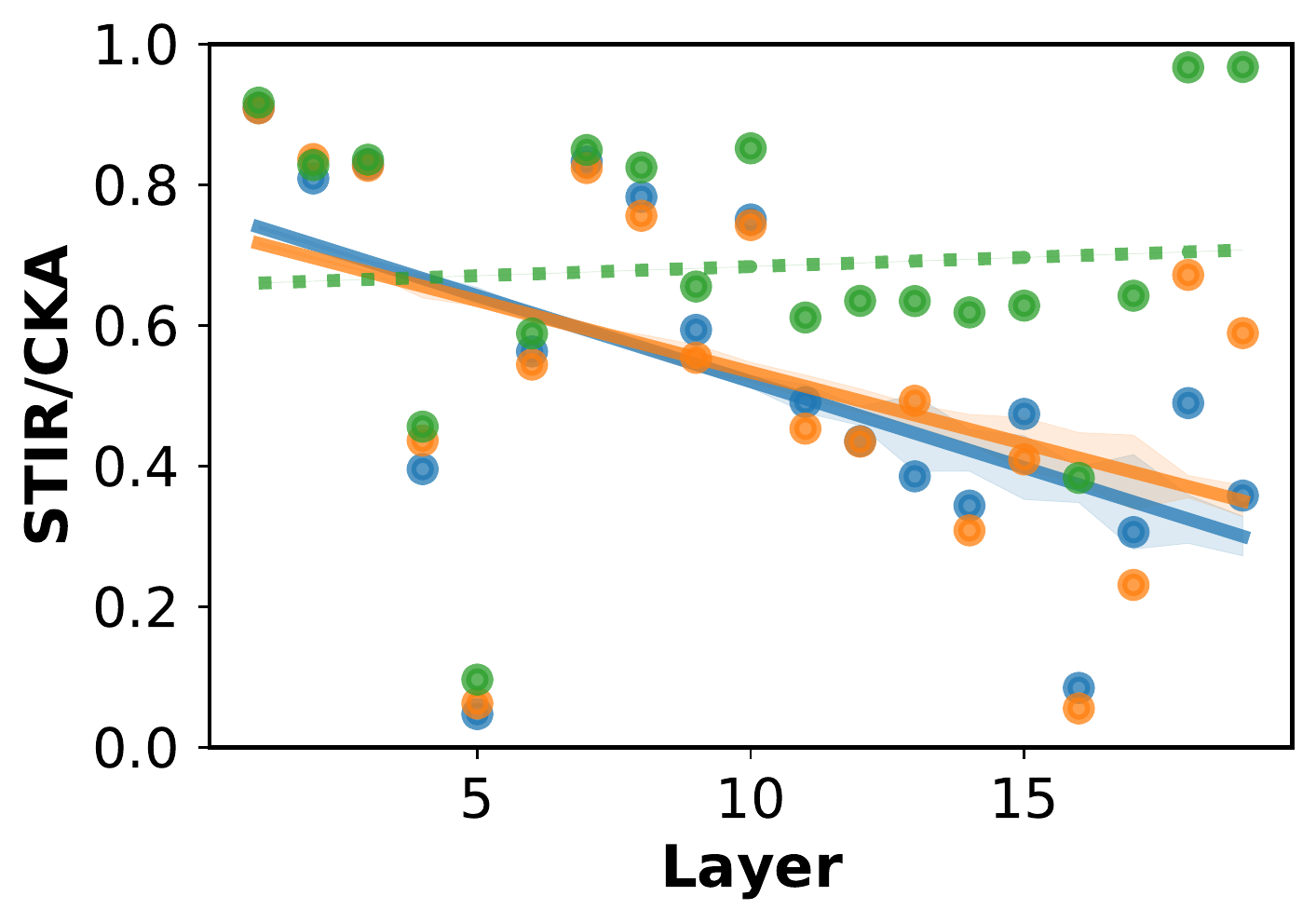}}
    \subfloat[2 ResNet18 with different init, AT]{\label{fig:rand_init_r18rob}
        \includegraphics[width=0.3\linewidth]{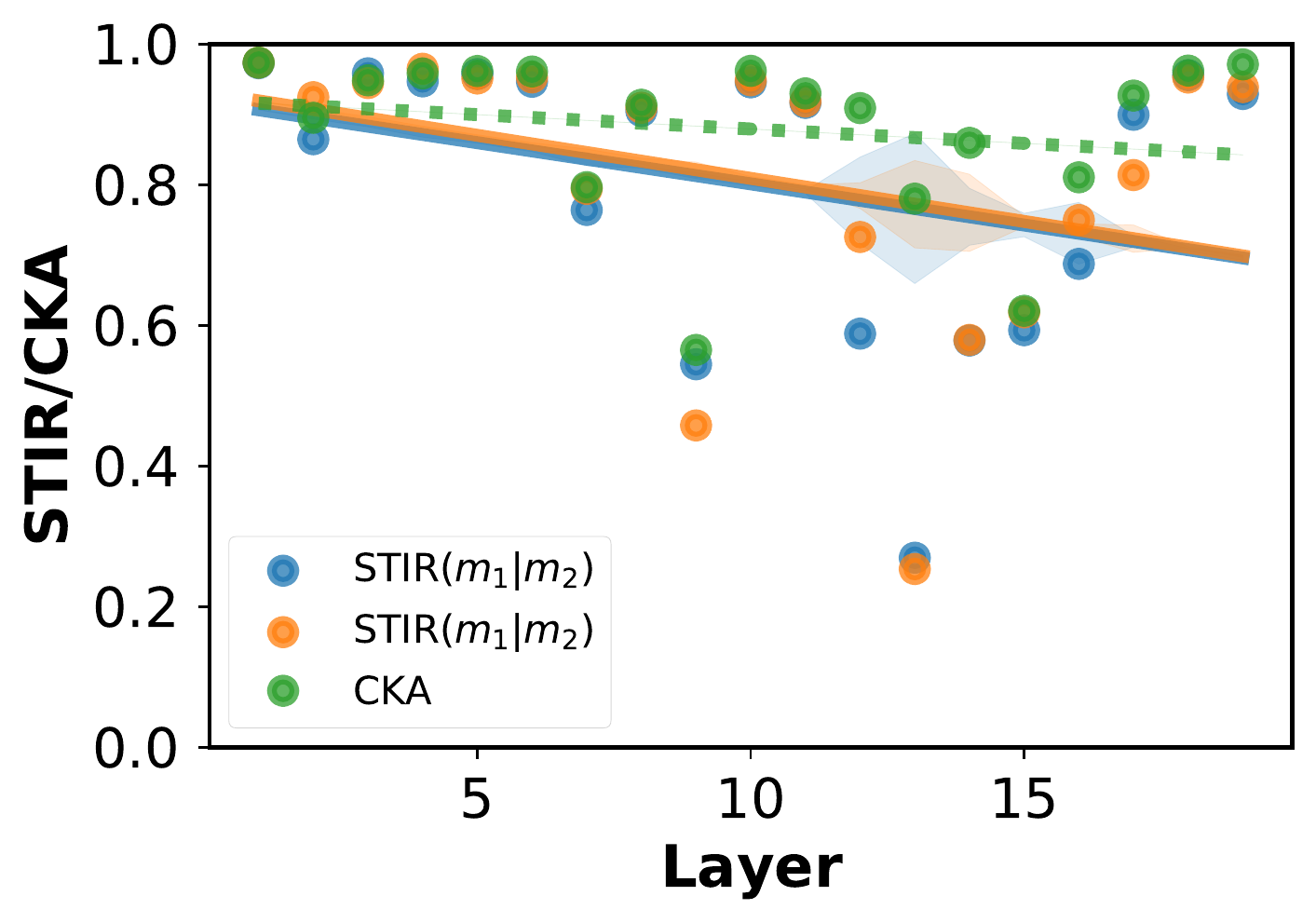}}
    
    \subfloat[2 VGG16 with different init, Vanilla]{\label{fig:rand_init_vgg16nonrob}
        \includegraphics[width=0.3\linewidth]{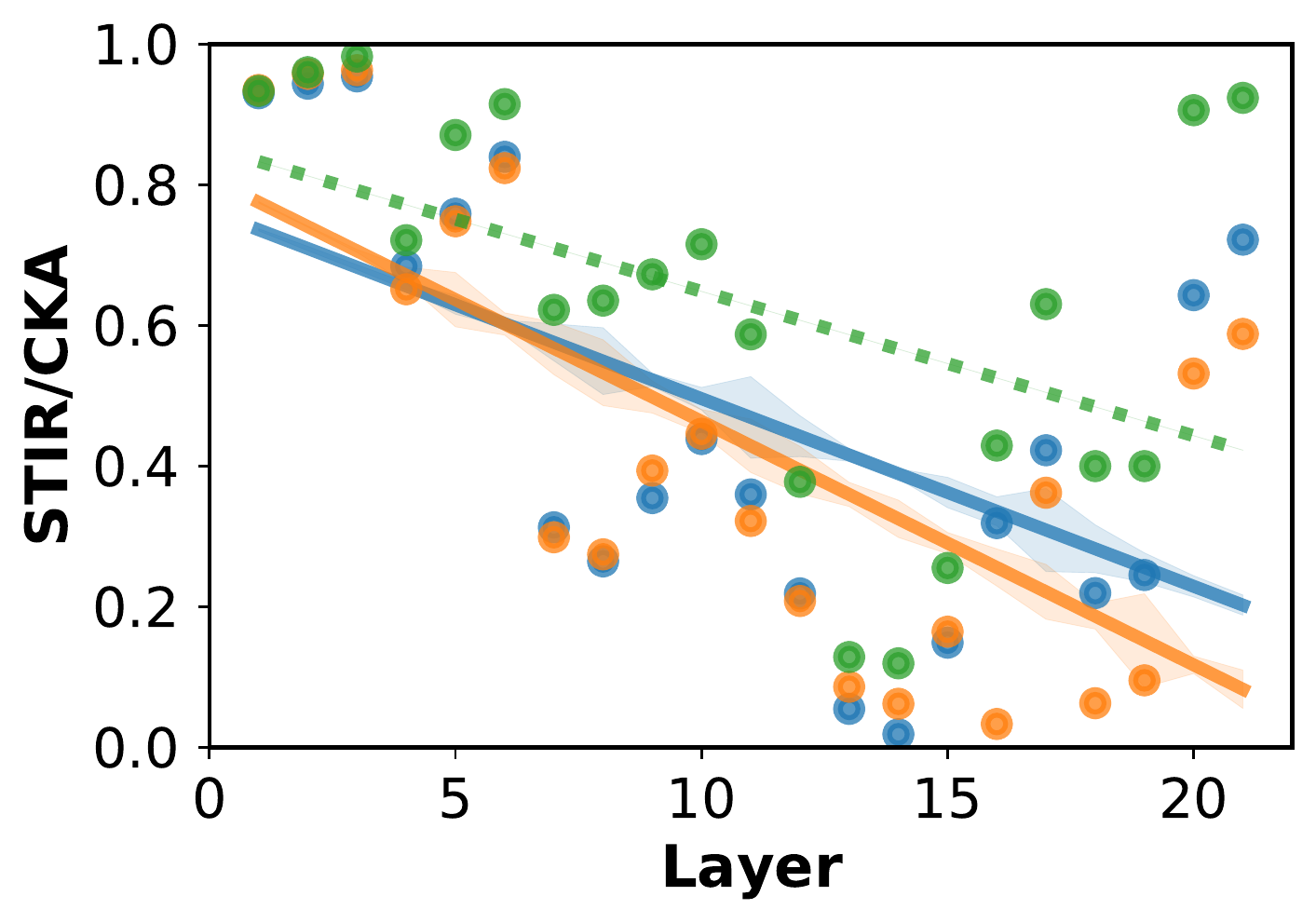}}
    \subfloat[2 VGG16 with different init, AT]{\label{fig:rand_init_vgg16rob}
        \includegraphics[width=0.3\linewidth]{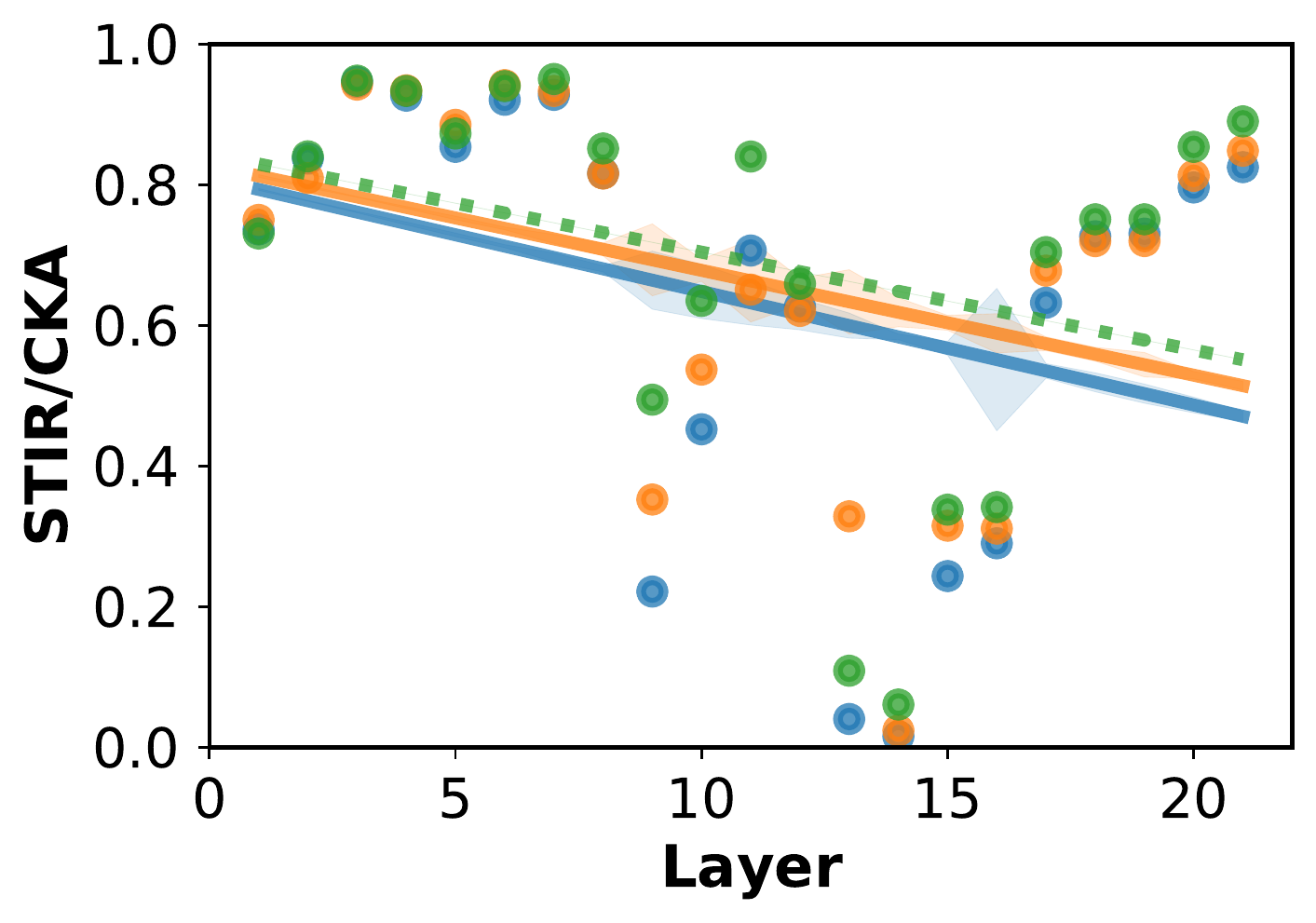}}
\caption{\textbf{[Role of Random Initialization; CIFAR10]} For models trained using the Vanilla crossentropy loss, we find that only initial layers have high shared invariances. However, when the training procedure explicitly introduces invariances (\eg adversarial training), then all layers converge to having similarly high shared invariances. Similar trends also hold for deeper variants of ResNet and VGG (results in Appendix~\ref{sec:appendix_insights}).}
\label{fig:rand_init}
\vspace{-4mm}
\end{figure*}

\begin{figure}[h!]
    \centering
    
    \subfloat[\mone = ResNet18 on CIFAR10 w Vanilla Loss\\\mtwo = ResNet18 on CIFAR100 w Vanilla Loss]{\label{fig:diff_datasets_r18erm}
        \includegraphics[width=0.7\linewidth]{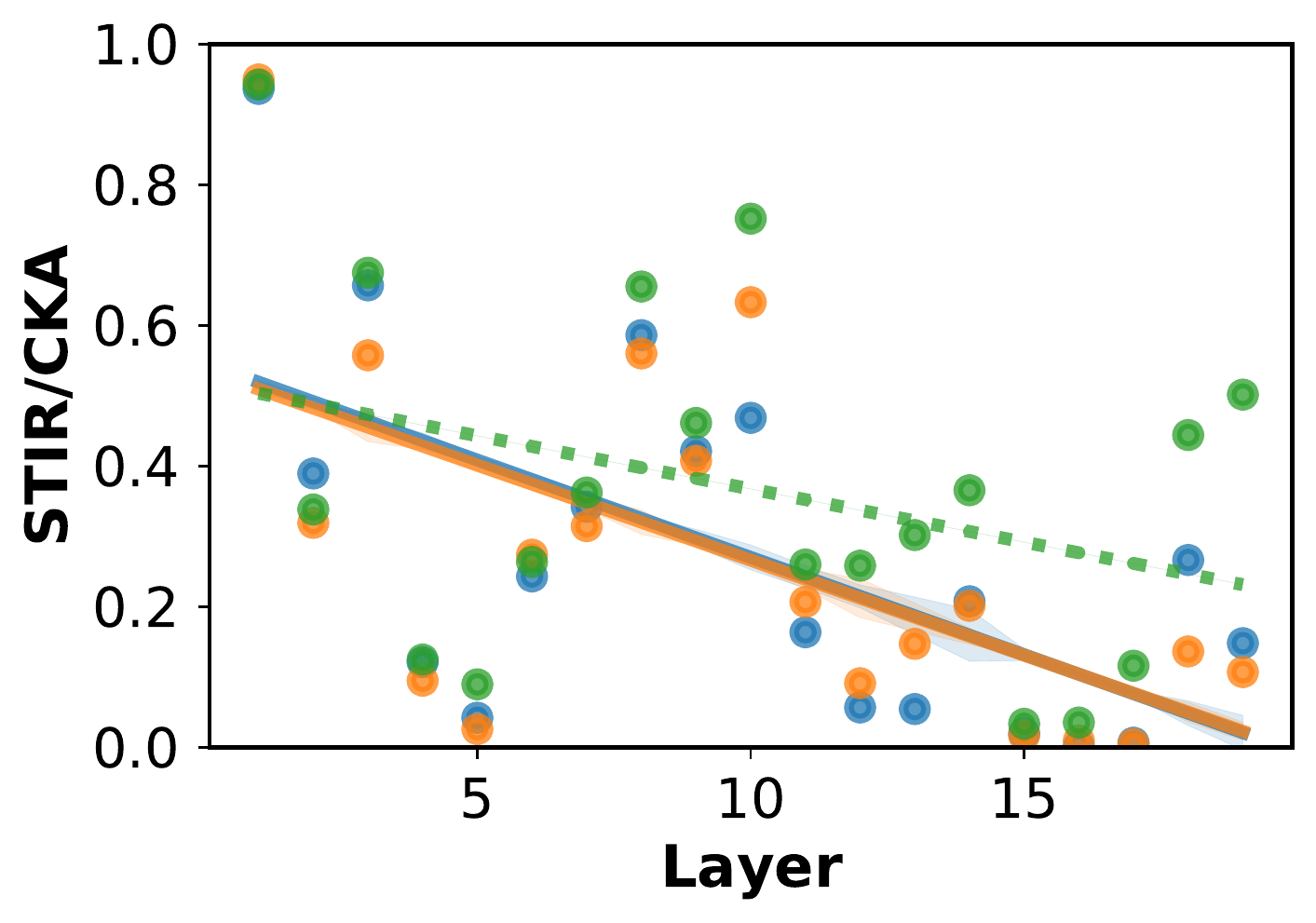}}
    
    \subfloat[\mone = ResNet18 on CIFAR10 w AT\\\mtwo = ResNet18 on CIFAR100 w AT]{\label{fig:diff_datasets_r18at}
        \includegraphics[width=0.7\linewidth]{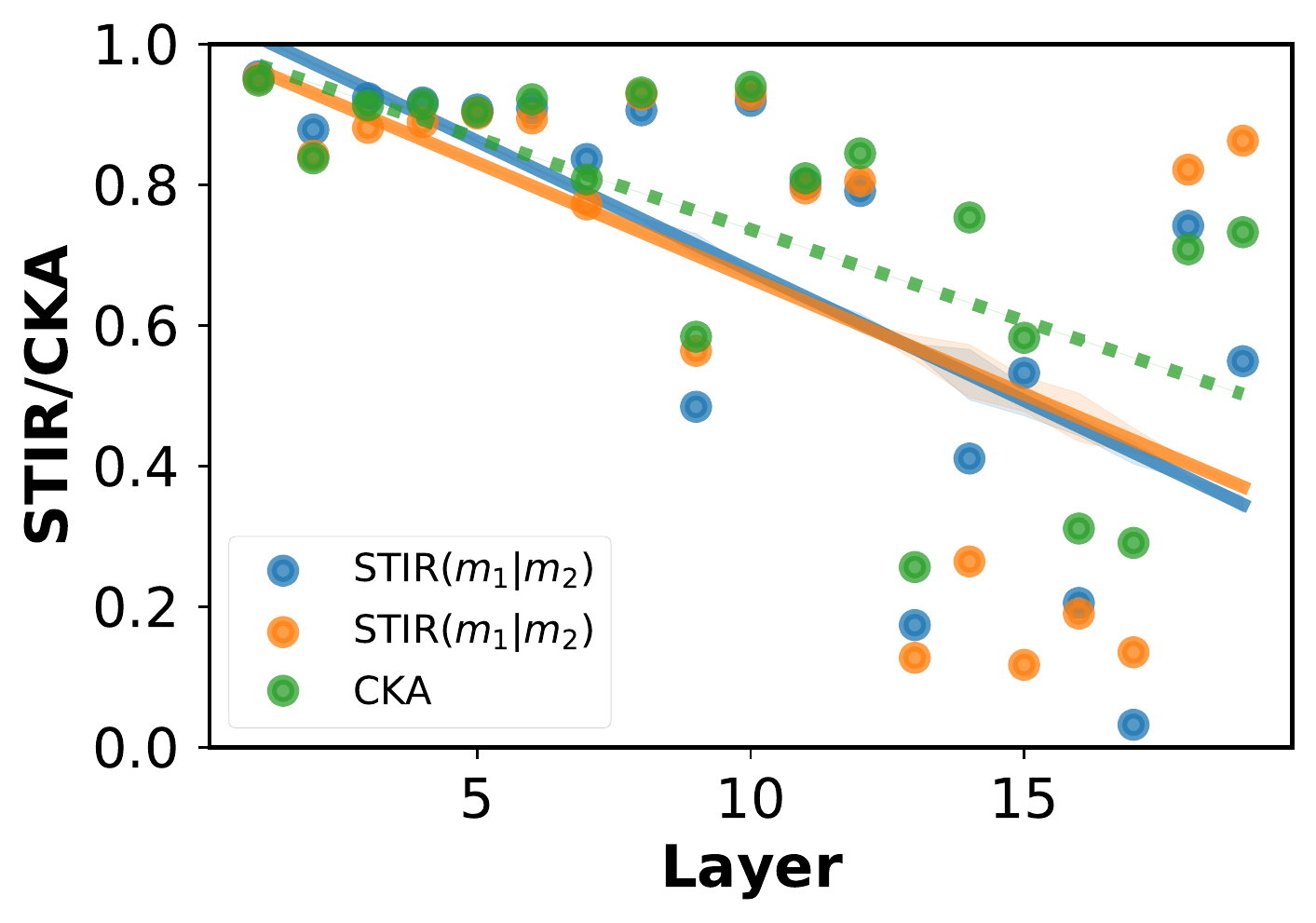}}

\caption{\textbf{[Different Datasets; ResNet18 on CIFAR10 and CIFAR100]} Similar to the finding of~\cite{kornblith2019similarity} we find that across datasets, initial layers tend to have more shared invariances. However, we also find that shared invariances increases substantially for all layers with adversarial training (AT).}
\label{fig:diff_datasets}
\end{figure}

\begin{figure}[h!]
    \centering
    
    \includegraphics[width=0.9\linewidth]{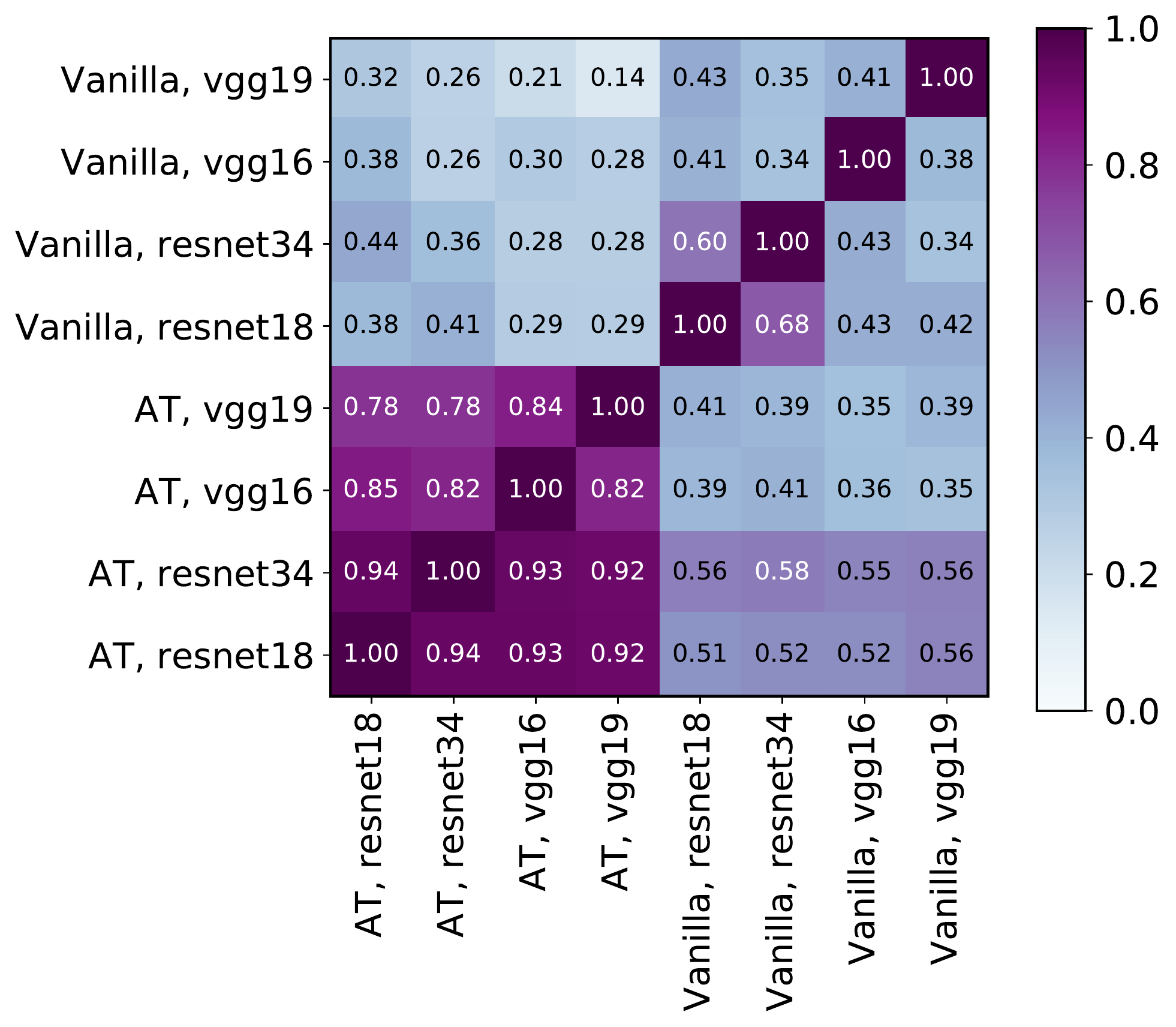}
\vspace{-4mm}
\caption{\textbf{[Different Architectures, Penultimate Layer Shared Invariances; CIFAR10]} We find that in general ResNets have high shared invariances when trained using the same loss, but this shared invariance drops across training types. We also see that with AT, even models with different architectures converge to high shared invariances. }
\label{fig:across_models_stir}
\vspace{-4mm}
\end{figure}

\begin{figure*}[h!]
    \centering
    
    \subfloat[AT (m$_1$) TRADES (m$_2$)]{\label{fig:diff_adv_r18madrytrades}
        \includegraphics[width=0.3\linewidth]{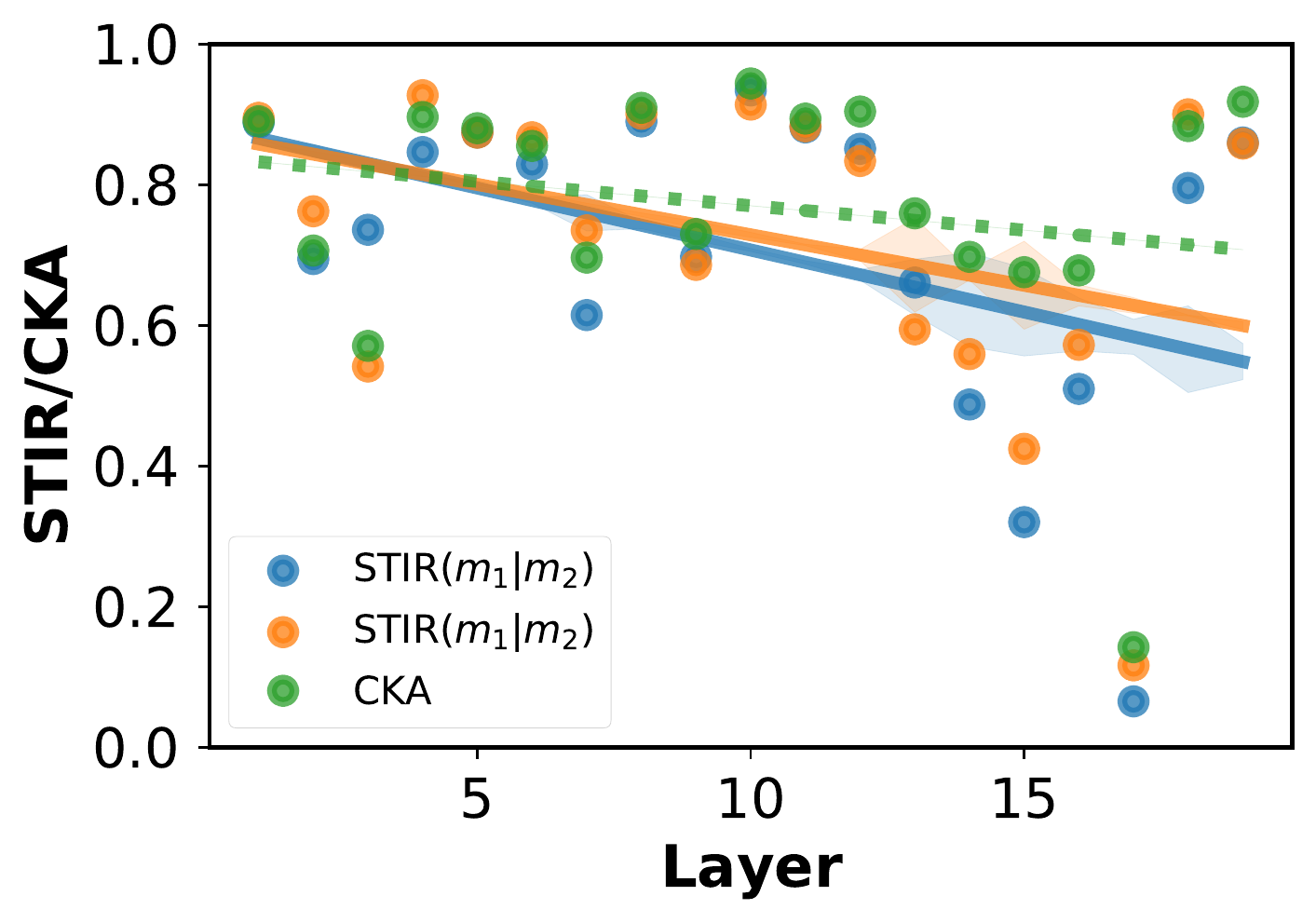}}
    \subfloat[AT (m$_1$) MART (m$_2$)]{\label{fig:diff_adv_r18madrymart}
        \includegraphics[width=0.3\linewidth]{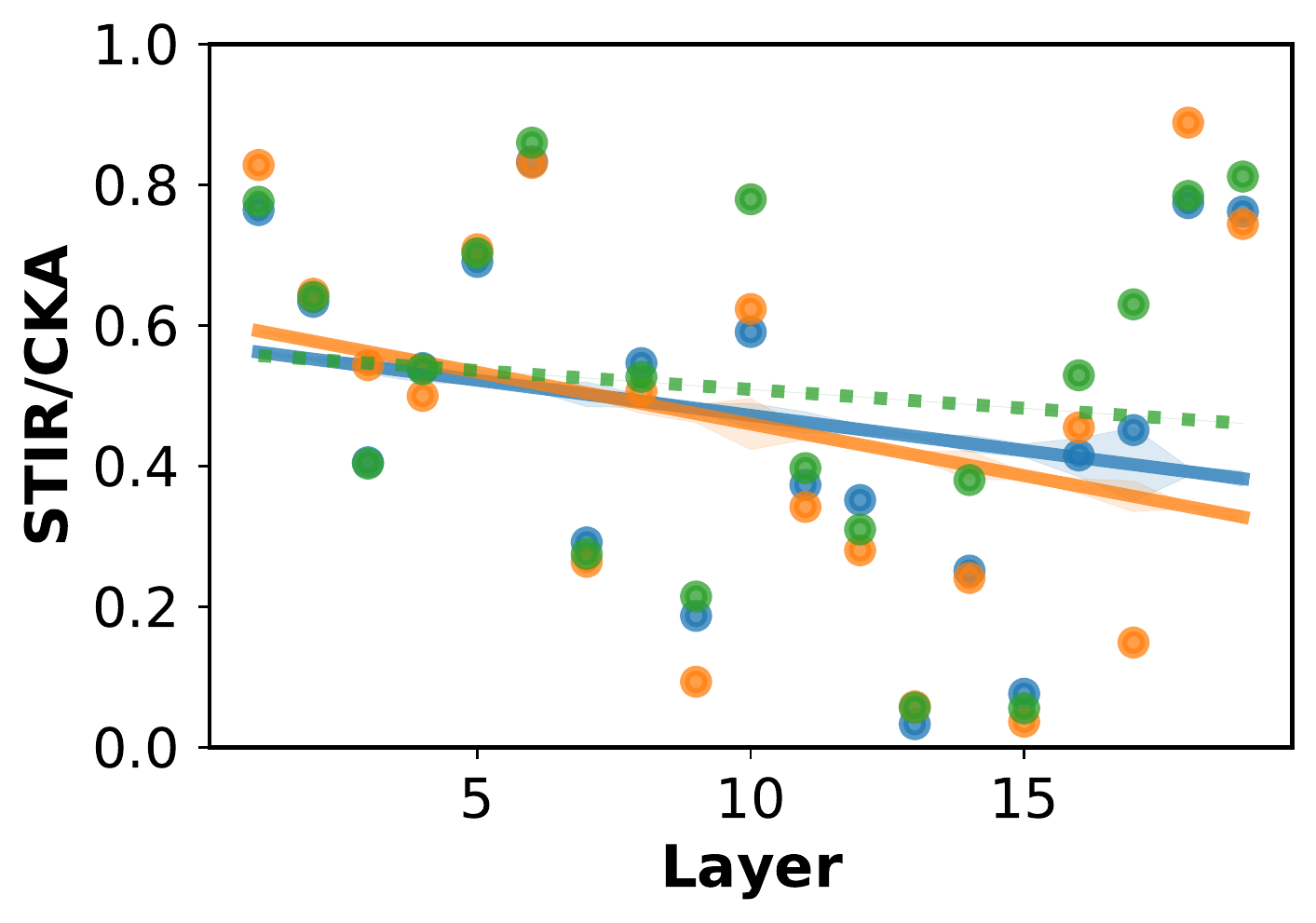}}
    \subfloat[TRADES (m$_1$) vs MART (m$_2$)]{\label{fig:diff_adv_r18tradesmart}
        \includegraphics[width=0.3\linewidth]{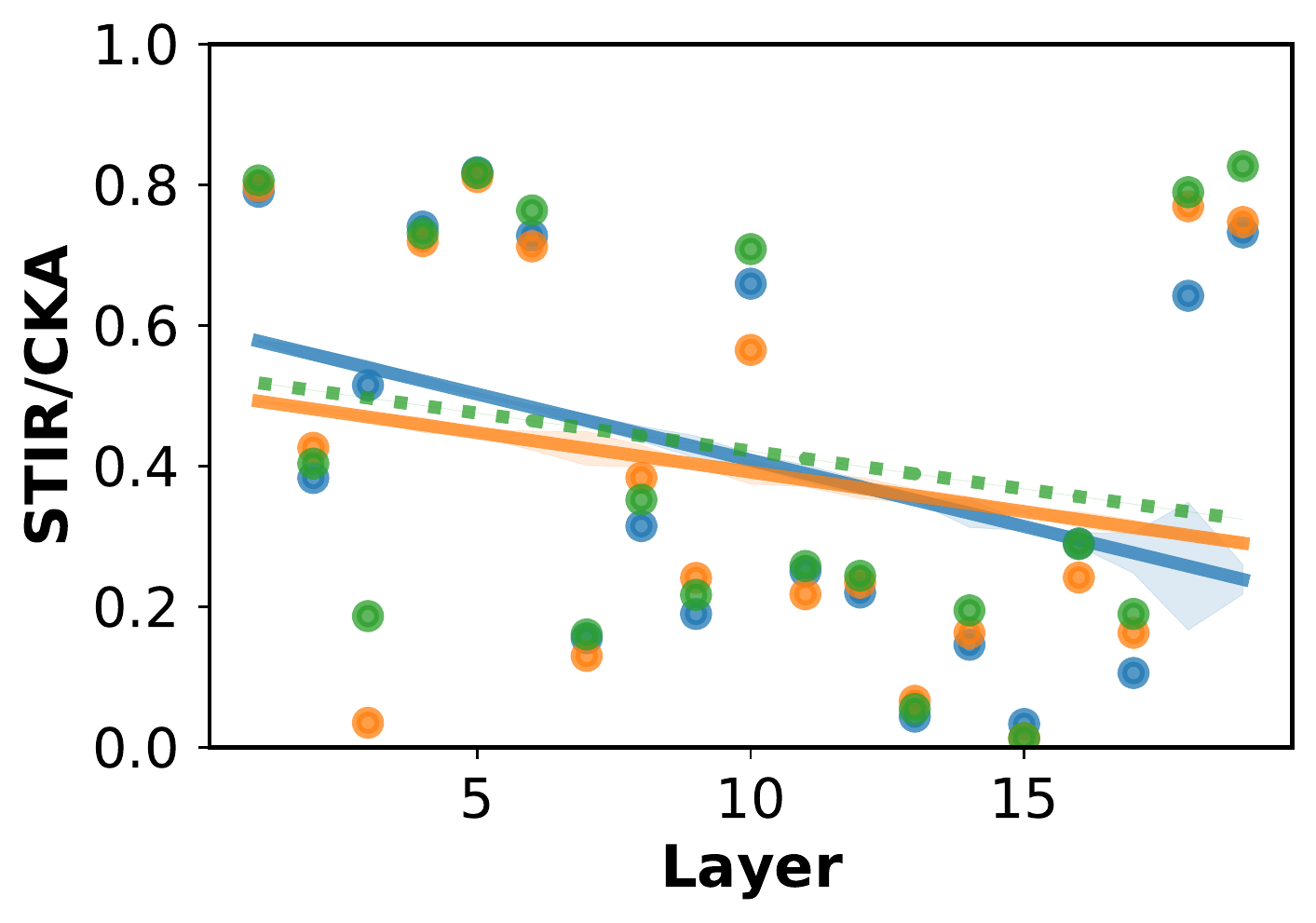}}
    
\caption{\textbf{[Different Types of Adversarially Robust Training; ResNet18 on CIFAR10]} Comparing same model trained using 3 different adversarial training variants: AT~\cite{madry2019deep}, TRADES~\cite{zhang2019theoretically} and MART~\cite{wang2019improving}. While these methods are geared towards the same goal of achieving invariance to $\ell_p$ perturbations, we find that other than TRADES and AT, these methods in fact do not have very high shared invariance (as opposed to similarity between two ResNet18s trained using AT, which have much higher values of shared invariance as shown in Fig.~\ref{fig:rand_init_r18rob}). This shows that these methods achieve resistance to $\ell_p$ ball attacks in very different ways.}
\label{fig:diff_adv}
\vspace{-4mm}
\end{figure*}

\section{Evaluating the Impact of Design Choices on Shared Invariance using \stir}

A lot of research effort been dedicated towards finding architectures, training schemes and datasets that produce more correct (\ie~accurate) models.
However, the effect that these design choices have on the relative invariance of models is still not properly understood.
In this section, we leverage \stir to investigate the effect that the various choices in the training pipeline have on shared invariances between models.
All evaluations of \stir in this section are performed using the CIFAR10 dataset. See Appendix~\ref{sec:appendix_exp_setup_details} for additional details.

\subsection{Role of Different Random Initialization, Different Training Datasets}

\textbf{Random Initializations} \cite{kornblith2019better} find that the same architecture trained from two different random initializations should converge to highly similar representations, \ie layer $k$ in both models has high similarity. We find that this is not necessarily the case for shared invariances. Fig.~\ref{fig:rand_init} shows results for two ResNet18 (different random initialization) and two VGG16 (different random initialization) models trained on CIFAR10 using the vanilla crossentropy loss and adversarial training. We see that models trained with vanilla loss (Fig.~\ref{fig:rand_init_r18nonrob} \&~\ref{fig:rand_init_vgg16nonrob}), later layers have lower shared invariances than initial layers, indicated by a negative slope of lines of best fit. A similar result showing high variance between two classifiers trained with vanilla loss that differ only in their random initialization has been observed for NLP models as well~\cite{zafar2021lack}, albeit for post-hoc explanations. This suggests a possibly interesting connection between shared invariance and post-hoc explanations which we leave for future work.
However, with adversarial training, both initial and final layers converge to (almost) similarly high levels of shared invariances (indicated by flatter lines of fit in Fig.~\ref{fig:rand_init_r18rob} \&~\ref{fig:rand_init_vgg16rob}). Thus, we conclude that when training from different random initializations, training procedures that explicitly introduce invariance (such as adversarial training) make each layer of two differently initialized models converge to similar shared invariances. 

\textbf{Datasets} For the same architecture (ResNet18) trained on different datasets (CIFAR10 and CIFAR100), we evaluate the shared invariances between each of the corresponding layers. We find that in general, initial layers tend to have higher shared invariances, as indicated by the negative slope of the line of best fit in Fig.~\ref{fig:diff_datasets}. Interestingly,~\cite{kornblith2019similarity} also had a similar observation when measuring similarity for models trained on different datasets. Additionally, we see that shared invariance increases substantially (for all layers) when these models are trained using adversarial training, similar to the random initilaization case, even though the training here is performed on different datasets. We see similar trends for other architectures too (results in Appendix~\ref{sec:appendix_insights}).

\subsection{Different Architectures, Penultimate Layer}

We train ResNet18, ResNet34, VGG16 and VGG19 with two different random seeds and using both the vanilla loss and adversarial training (AT). We then compute the shared invariances across all these configurations of models (for the penultimate layer of each model) as shown in Fig.~\ref{fig:across_models_stir}. 

We find that in general, architectures with residual connections (ResNet18 and ResNet34) have high shared invariances amongst themselves, as indicated by high values of \stir amongst ResNets (for both vanilla and AT). 

\subsection{Different Adversarial Training Methods}

As observed in Fig.~\ref{fig:across_models_stir}, models trained with AT generally have higher values of \stir amongst them than models trained using the vanilla loss. This raises a natural question: does high \stir between models hold for any kind of training that makes models robust to $\ell_p$ ball perturbations?

To test this, we train a ResNet18 on CIFAR10 using 3 distinct training losses, all of which have been shown to be highly robust to $\ell_p$ ball perturbations: Adversarial Training (AT)~\cite{madry2019deep}, TRADES~\cite{zhang2019theoretically} and MART~\cite{wang2019improving}. TRADES and MART contains an additional hyperparameter $\beta$ that is used to trade off between accuracy and adversarial robustness. We use $\beta = 0.1$ for our experiments, and additional details can be found in Appendix~\ref{sec:appendix_exp_setup_details}. All 3 of these ResNets achieve similar clean and robust accuracy.

Surprisingly, we find that models trained using these methods only have mild levels of shared invariance, as indicated by lines of best fit around $0.5$ (see Fig.~\ref{fig:diff_adv}). This is in contrast to the case of two ResNet18s trained using AT (see Fig.~\ref{fig:rand_init_r18rob}) which leads to very high shared invariance. This shows that the differences shown in Fig.~\ref{fig:diff_adv} are not due to stochasticity in training but rather due to the training methods themselves. Thus, while all of these methods achieve the same goal of robustness in an $\ell_p$ ball, they achieve so in very different ways. We see similar trends for other architectures too (results in Appendix~\ref{sec:appendix_insights}).

In summary, we find that shared invariance between models tends to decrease with increasing layer-depth and adversarial training significantly increases the degree of shared invariance.
Models with architectures using residual connections exhibit a higher degree of shared invariance, whereas different methods of adversarial training do not necessarily lead to models with the same shared invariances. %
\section{Related Work}
\label{sec:rel_work}

\xhdr{Comparing Representations.}
A number of papers have proposed methods for measuring the similarity of representations in deep neural networks \cite{laakso2000content,li2016convergent,wang2018towards,raghu2017svcca,morcos2018insights,kornblith2019similarity}.
Our invariance measure leverages CKA \cite{kornblith2019similarity}, however, we discuss in Sections \ref{sec:sim_problems_logical} and \ref{sec:sim_problems_empirical} why the existing metrics themselves are unsuitable for measuring shared invariances between models.
Recent work has identified lack of consistency between existing similarity measures and proposed a measure that is consistent~\cite{ding2021grounding}. However, their proposed measure also measures similarity and does not take into account the invariances. It can thus be used in conjunction with our proposed measure, but not replace it for measuring invariance. There has been work on measuring shared invariance between NN representations and (black box) humans~\cite{feather2019metamers,nanda2022exploring}, however, we consider a different problem of shared invariances between two NNs.

\xhdr{Understanding Representations.}
A lot of work on scrutinizing representations has been geared towards improving the interpretability of neural networks~\cite{mahendran2014understanding,olah2017feature,kim2018interpretability,dosovitskiy2016generating,dosovitskiy2016inverting}. We're particularly inspired by representation inversion~\cite{mahendran2014understanding,dosovitskiy2016inverting} which is a key component of our proposed measure.
However, while the goals of all of these works is to be able to make a neural network more interpretable to humans, \ie~to enable qualitative judgements about the network's behavior, our goal is to instead measure the degree of shared invariance between any two models, \ie~to make quantitative statements. Recent work has used model stitching to compare representations~\cite{bansal2021revisiting}, but it is focused on better understanding of learned representations, rather than measuring robustness.
\citet{higgins18towards} define disentanglement in representation learning by leveraging the concept of invariance. Their work, however, only provides a definition of disentanglement and no measure for the degree of invariance.

\xhdr{Robustness.} Our work takes inspiration from many works on adversarial~\cite{szegedy2013intriguing,madry2019deep,zhang2018perceptual,ilyas2019adversarial}, natural~\cite{hendrycks2018benchmarking,geirhos2018generalisation} and distributional~\cite{taori2020measuring,recht2019imagenet,koh2021wilds} robustness. However, in all these works, the reference model is (implicitly) assumed to be a human and the goal is to make a neural network follow the invariances of a human. In our work we explicitly characterize the reference model, which can be another neural network, that allows us to unify all the different notions of robustness.

\section{Conclusion}
We proposed a directional measure of shared invariance between representations that takes into account the invariances of the model that generated these representations. We showed how our measure faithfully estimates shared invariances where existing representation similarity methods may fail. Furthermore, we showed how our measure can be used to derive interesting insights about deep learning. It will be interesting to explore this direction further in future work, revisiting earlier analysis based on previous representation similarity approaches \cite{nguyen2020wide,raghu2021vision}.
Another interesting avenue to explore is how our measure can be used during training to encourage one model to follow similar invariances to another. This could be helpful to update a neural network in safety-critical applications by ensuring that the new network maintains the invariances of the original model. 

\section*{Acknowledgements}
AW acknowledges support from a Turing AI Fellowship under grant EP/V025279/1, The Alan Turing Institute, and the Leverhulme Trust via CFI. VN, TS, and KPG were supported in part by an ERC Advanced Grant “Foundations for Fair Social Computing” (no. 789373). VN and JPD were supported in part by NSF CAREER Award IIS-1846237, NSF D-ISN Award \#2039862, NSF Award CCF-1852352, NIH R01 Award NLM013039-01, NIST MSE Award \#20126334, DARPA GARD \#HR00112020007, DoD WHS Award \#HQ003420F0035, ARPA-E Award \#4334192 and a Google Faculty Research Award. Finally, we would like to thank anonymous reviewers for their constructive feedback and suggestions for experiments in Section~\ref{sec:stir_uses}. %

\bibliography{relative_robustness}
\bibliographystyle{style_files/icml2022}

\appendix
\section{Representation Similarity Measures}
\label{sec:appendix_similarity_measures}

A recent line of work has studied measures for the similarity of representations in deep neural networks.
Given two models $m_1$ and $m_2$ and a set $Z$ of input points to both, all of these measures quantify the degree of representational similarity between two sets of representations $X = m_1(Z) \in \R^{n \times d_1}$ and $Y = m_2(Z) \in \R^{n \times d_2}$ as $\orsim(X, Y)$ and are defined as follows:

\textbf{\svcca} \cite{raghu2017svcca} is computed via the following steps:
\begin{enumerate}
    \item Compute $X', Y'$ by pruning $X, Y$ using SVD to only retain the first $k_1$ and $k_2$ principal components that are necessary to retain 99\% of the variance, respectively.
    \item Perform Canonical Correlation Analysis $\cca(X', Y')$ yielding $k = \min(k_1, k_2)$ correlation coefficients $\rho_1,\dots,\rho_k$.
    \item Return $\orsim(X, Y) = \frac{1}{k} \sum_{i=1}^k \rho_i$
\end{enumerate} 

\textbf{\pwcca} \cite{morcos2018insights} is computed via the following steps:
\begin{enumerate}
    \item Perform Canonical Correlation Analysis $\cca(X, Y)$ yielding $k = min(d_1, d_2)$ correlation coefficients $\rho_1,\dots,\rho_k$ and \cca vectors $h_1,\dots,h_k$
    \item Compute weights $\alpha_i = \sum_{j=1}^{d_1} |\langle h_i, z_j \rangle|$, where the $z_j$'s are the $d_1$ columns of $X$
    \item Return $\orsim(X, Y) = \frac{\sum_{i=1}^k \alpha_i \rho_i}{\sum_{i=1}^k \alpha_i}$
\end{enumerate}

\textbf{(Linear) \cka} \cite{kornblith2019similarity} is computed via the following steps:
\begin{enumerate}
    \item Compute similarity matrices $K = X X^T$, $L = Y Y^T$
    \item Compute normalized versions $K' = H K H$, $L' = H L H$ of the similarity matrix using centering matrix $H = I_n - \frac 1 n \1 \1^T$
    \item Return $\cka(X, Y) = \frac{\text{HSIC}(K, L)}{\sqrt{\text{HSIC}(K, K) \text{HSIC}(L, L)}}$, where $\text{HSIC}(K, L) = \frac{1}{(n-1)^2} \text{flatten}(K') \cdot \text{flatten}(L')$
\end{enumerate}

\section{Model Architecture, Hardware, Training, and Other Details}\label{sec:appendix_exp_setup_details}

We use ResNet18, ResNet34~\cite{he2016deep}, VGG16 and VGG19~\cite{simonyan2014very} trained on CIFAR10/100~\cite{krizhevsky2009learning} using the standard crossentropy loss and other adversarially robust training methods such as AT, TRADES and MART. All of our experiments are performed on standard models and datasets that can fit on standard GPUs. We also attach our code to reproduce the numbers. For all purposes of adversarial training we use the $\ell_2$ threat model with $\epsilon = 1.0$ (see~\cite{madry2019deep} for details). Additionally TRADES and MART require another hyperparameter $\beta$ (that balances adversarial robustness and clean accuracy) which we set to $1.0$ for our experiments. 

\section{\stir Faithfully Measures Shared Invariance}\label{sec:appendix_faithful_stir}

To confirm that \stir is correct in assigning different scores to two (same) models trained from different initializations, we generate controversial stimuli~\cite{golan2020controversial} for all configurations in Table~\ref{tab:walkthorugh_intuitive_comparison}: ($\merm_1, \merm_2$), ($\madv_1, \madv_2$), and ($\merm_1, \madv_2$). Such stimuli are generated by solving the following optimization for a training data point $x$ and any two models \mone and \mtwo:

\begin{equation}\label{eq:controversial_stimuli}
    \text{argmin}_{x_c} \,\, || m_1(x) - m_1(x_c) ||_2 - || m_2(x) - m_2(x_c) ||_2 
\end{equation}

Here we assume $m_1(.)$ and $m_2(.)$ are penultimate layer representations of the respective models. This process generates $x_c$ for every point $x$ such that $|| m_1(x) - m_1(x_c) ||_2$ is low and $|| m_2(x) - m_2(x_c) ||_2$ is high. Since these $x_c$ are ``perceived'' very differently by the two models (wrt the original $x$), they are called \textit{controversial stimuli}. We use the empirical mean of the amount of perturbation needed (\ie $\delta = \E_{x_c, x} [||x - x_c||_2]$) as a measure of ease of generating controversial stimuli~\footnote{High(er) values of $\delta$ indicate it was hard(er) to generate controversial stimuli}. For a pair of models (\mone, \mtwo) where it's easy to generate such $x_c$, the shared invariance should be low. We see that this is indeed the case as indicated by numbers in Table~\ref{tab:walkthorugh_intuitive_comparison}.

\def\arraystretch{1.25}
\begin{table*}[!t]
\begin{center}
\begin{small}
\begin{sc}
\begin{tabular}{c|cc|cc|cc}
\hline

& \multicolumn{2}{c|}{\begin{tabular}[c]{@{}c@{}}ResNet18 Vanilla (\textbf{$m_1$}),\\ResNet18 Vanilla (\textbf{$m_2$}) \end{tabular}} & \multicolumn{2}{c|}{\begin{tabular}[c]{@{}c@{}}ResNet18 AT (\textbf{$m_1$}),\\ResNet18 AT (\textbf{$m_2$}) \end{tabular}} & \multicolumn{2}{c}{\begin{tabular}[c]{@{}c@{}}ResNet18 AT (\textbf{$m_1$}),\\ResNet18 Vanilla (\textbf{$m_2$}) \end{tabular}} \\
 & & & & & \\
\hline

 & \multirow{2}{*}{\begin{tabular}[c]{@{}c@{}}\bf{$m_1 | m_2$}\end{tabular}} & \multirow{2}{*}{\begin{tabular}[c]{@{}c@{}}\bf{$m_2 | m_1$}\end{tabular}} &  \multirow{2}{*}{\begin{tabular}[c]{@{}c@{}}\bf{$m_1 | m_2$}\end{tabular}} & \multirow{2}{*}{\begin{tabular}[c]{@{}c@{}}\bf{$m_2 | m_1$}\end{tabular}} &  \multirow{2}{*}{\begin{tabular}[c]{@{}c@{}}\bf{$m_1 | m_2$}\end{tabular}} & \multirow{2}{*}{\begin{tabular}[c]{@{}c@{}}\bf{$m_2 | m_1$}\end{tabular}} \\ 
 & & & & & \\ \hline

\stir & $0.605_{\pm 0.013}$ & $0.562_{\pm 0.023}$ & $0.934_{\pm 0.003}$ & $0.939_{\pm 0.002}$ & $0.405_{\pm 0.020}$ & $0.509_{\pm 0.011}$ \\ 
$\ostir_{adv}$ & $0.085_{\pm 0.004}$ & $0.064_{\pm 0.004}$ & $0.096_{\pm 0.007}$ & $0.078_{\pm 0.005}$ & $0.054_{\pm 0.004}$ & $0.070_{\pm 0.004}$ \\ 
\cka & \multicolumn{2}{c|}{$0.967_{\pm 0.000}$} & \multicolumn{2}{c|}{$0.937_{\pm 0.000}$} & \multicolumn{2}{c}{$0.536_{\pm 0.000}$} \\
Acc($m_1(X')$, $m_2(X')$) & $0.521_{\pm 0.061}$ & $0.347_{\pm 0.029}$ & $0.891_{\pm 0.007}$ & $0.901_{\pm 0.002}$ & $0.140_{\pm 0.028}$ & $0.555_{\pm 0.012}$ \\
$\Delta \,\, (= \frac{1}{n}\sum|| X - X' ||_2)$ & $8.588_{\pm 0.491}$ & $8.570_{\pm 0.474}$ & $17.106_{\pm 3.204}$ & $16.210_{\pm 2.193}$ & $19.270_{\pm 2.202}$ & $7.368_{\pm 0.602}$ \\ \hline

\end{tabular}
\end{sc}
\end{small}
\end{center}

\caption{
\textbf{[\stir faithfully estimates shared invariance]} Here the two ResNet18s in each column are trained on CIFAR10 with different random initializations, holding every other hyperparameter constant. 1.) For two such models trained using the vanilla crossentropy loss (left), interestingly, we find that \stir highlights a lack of shared invariance, whereas \cka overestimates this value; 2.) when both models are trained using adversarial training~\cite{madry2019deep} (middle) \stir faithfully estimates high shared invariance; 3.) Finally \stir is able show how having a directional measure can bring out the differences when comparing a model trained with vanilla loss and adv training (right), whereas \cka being unidirectional cannot derive these insights. All numbers are computed over 1000 random samples from CIFAR10 training set and averaged over 5 runs. Additionally we show $\Delta$ between original inputs (X) and controversial stimuli (X') for a given configuration of $m_1$ and $m_2$. We see that controversial stimuli are ``hard'' to generate for higher STIR scores. Here hardness is indicated by the amount of $\ell_2$ perturbation needed to generate controversial stimuli.}
\label{tab:walkthorugh_intuitive_comparison}

\end{table*}

\section{Experiments: Insights using \stir}\label{sec:appendix_insights}

\subsection{Role of Different Random Initialization, Different Training Datasets}

\begin{figure*}[t!]
    \centering
    
    \subfloat[2 ResNet34 with different init, Vanilla]{\label{fig:rand_init_r34nonrob}
        \includegraphics[width=0.3\linewidth]{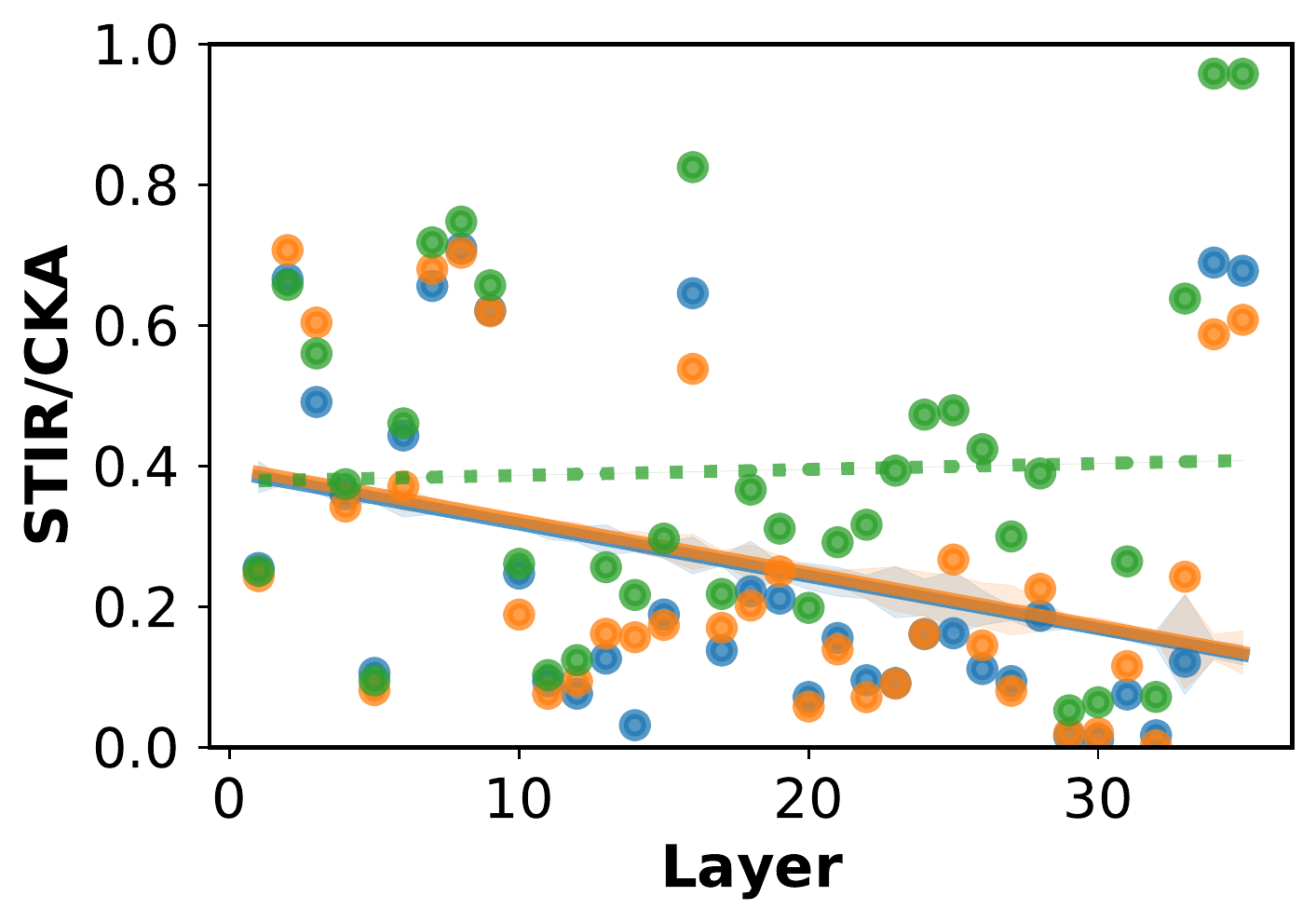}}
    \subfloat[2 ResNet34 with different init, AT]{\label{fig:rand_init_r34rob}
        \includegraphics[width=0.3\linewidth]{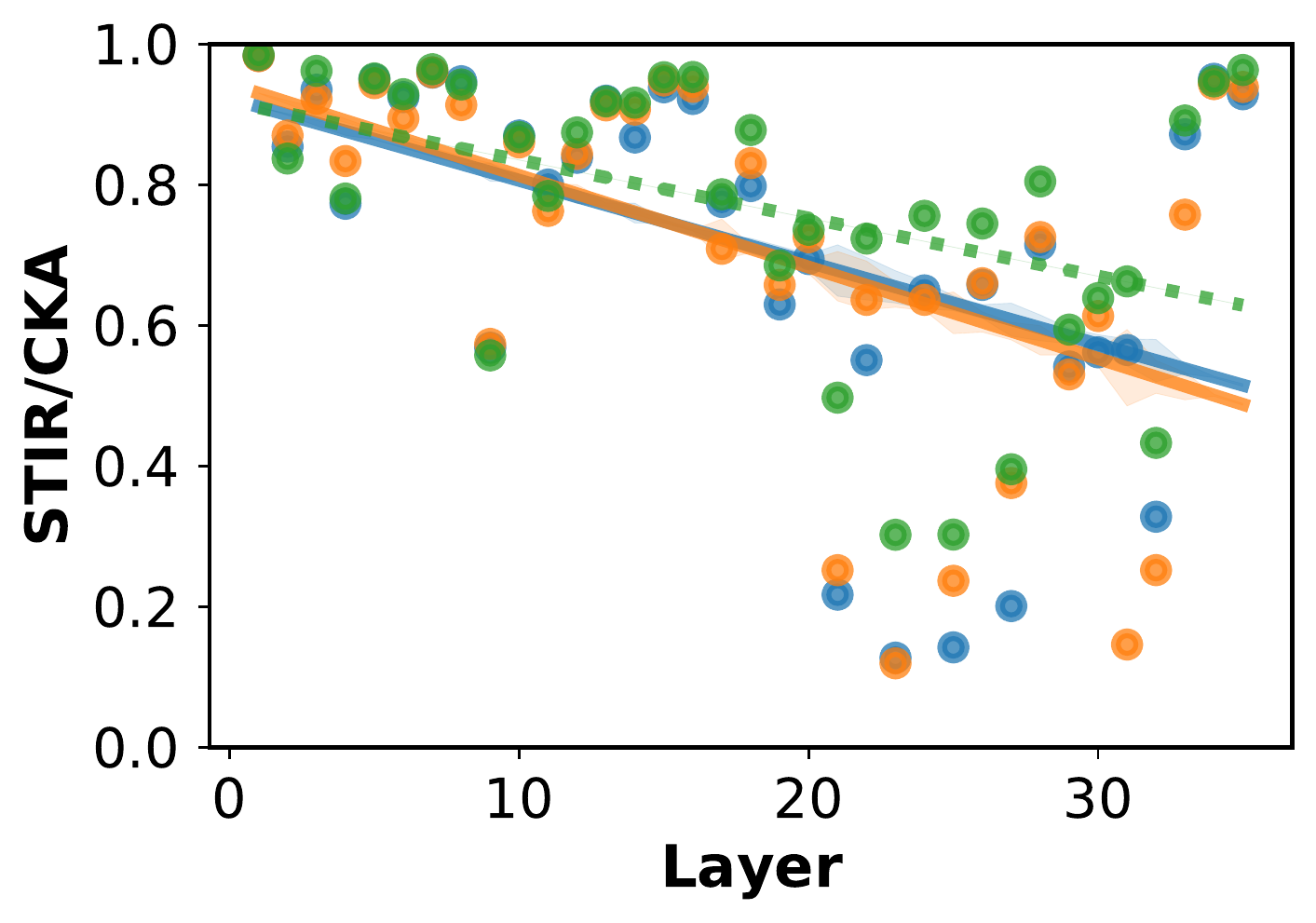}}
    
    \subfloat[2 VGG19 with different init, Vanilla]{\label{fig:rand_init_vgg19nonrob}
        \includegraphics[width=0.3\linewidth]{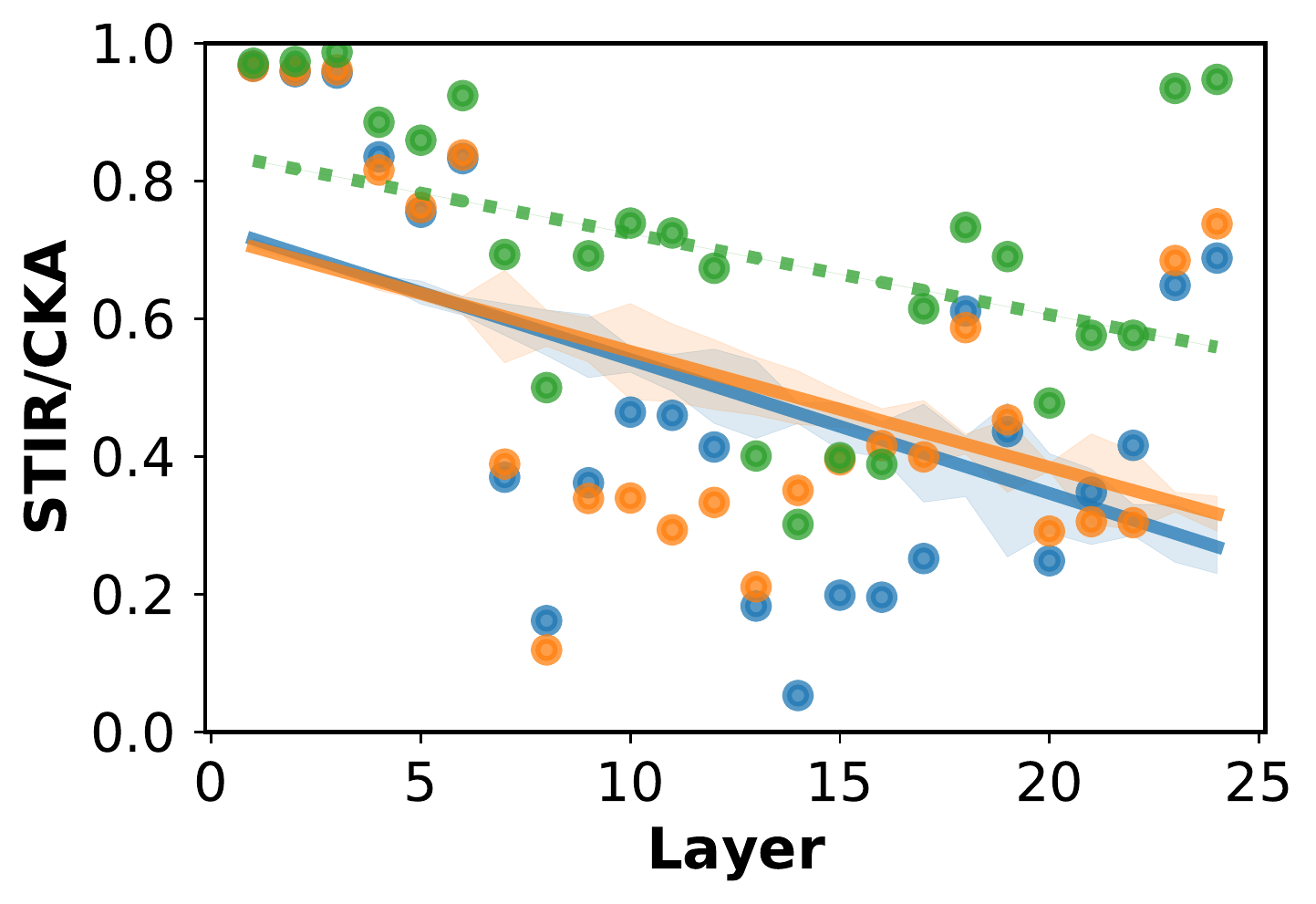}}
    \subfloat[2 VGG19 with different init, AT]{\label{fig:rand_init_vgg19rob}
        \includegraphics[width=0.3\linewidth]{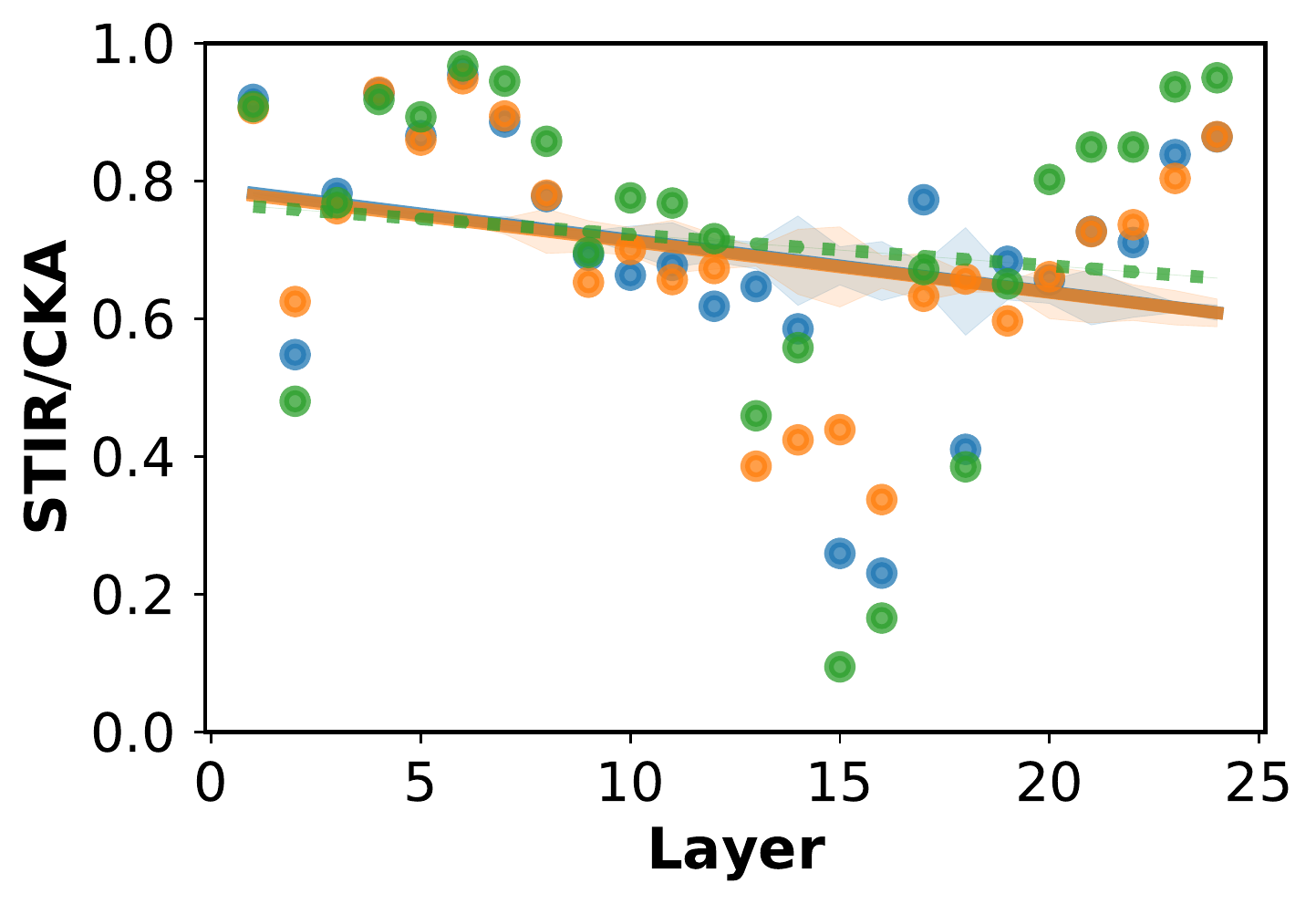}}
\caption{\textbf{[Role of Random Initialization; CIFAR10]} For models trained using the Vanilla crossentropy loss, we find that only initial layers have high shared invariances. However, when the training procedure explicitly introduces invariances (\eg adversarial training), then all layers converge to having similarly high shared invariances. We see similar trends for ResNet34 and VGG19 here.}
\label{fig:rand_init_more_models}
\end{figure*}

\begin{figure*}[h!]
    \centering
    
    \subfloat[\mone = ResNet34 on CIFAR10 w ERM\\\mtwo = ResNet34 on CIFAR100 w ERM]{\label{fig:diff_datasets_r34erm}
        \includegraphics[width=0.35\linewidth]{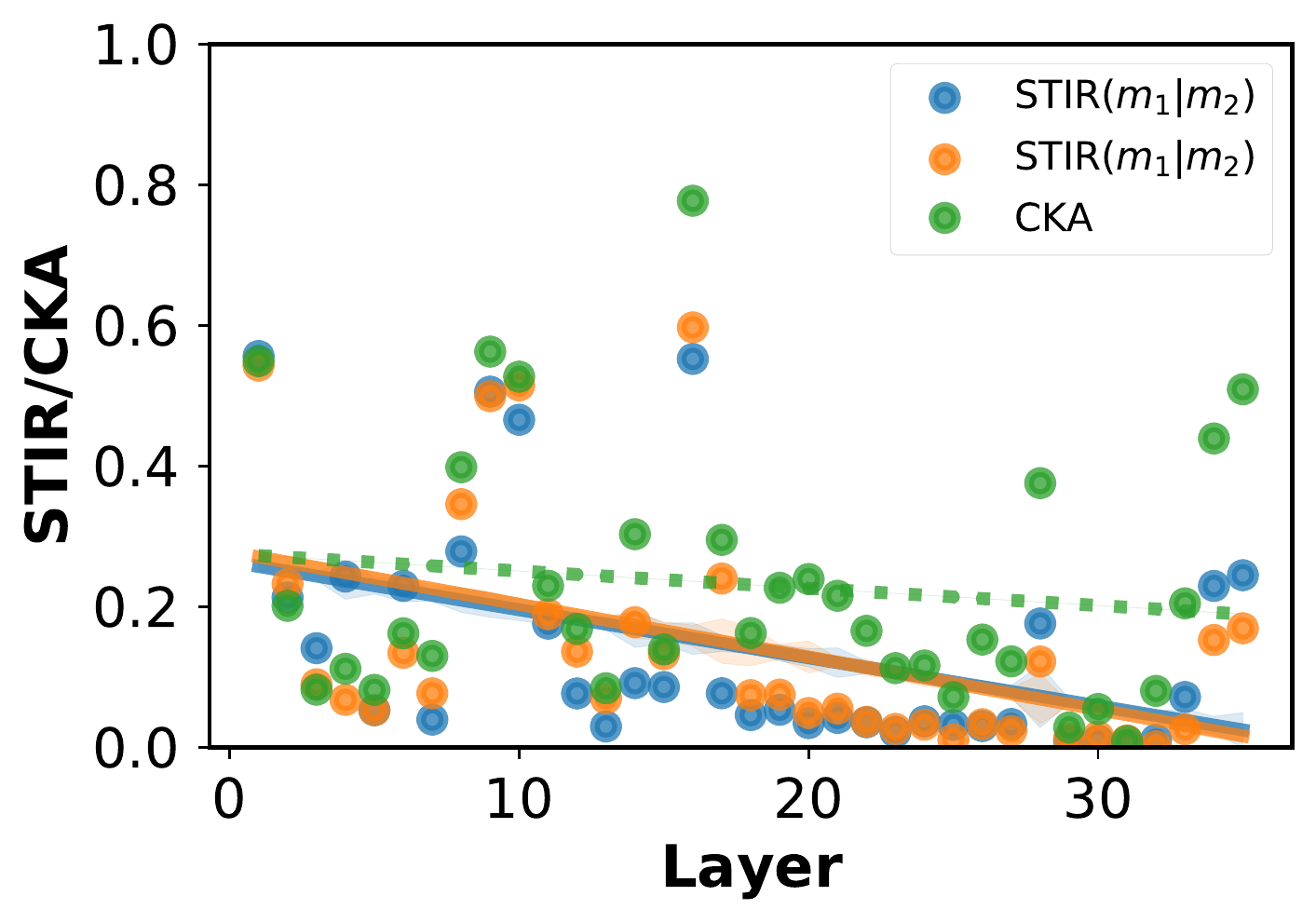}}
    \subfloat[\mone = ResNet34 on CIFAR10 w AT\\\mtwo = ResNet34 on CIFAR100 w AT]{\label{fig:diff_datasets_r34at}
        \includegraphics[width=0.35\linewidth]{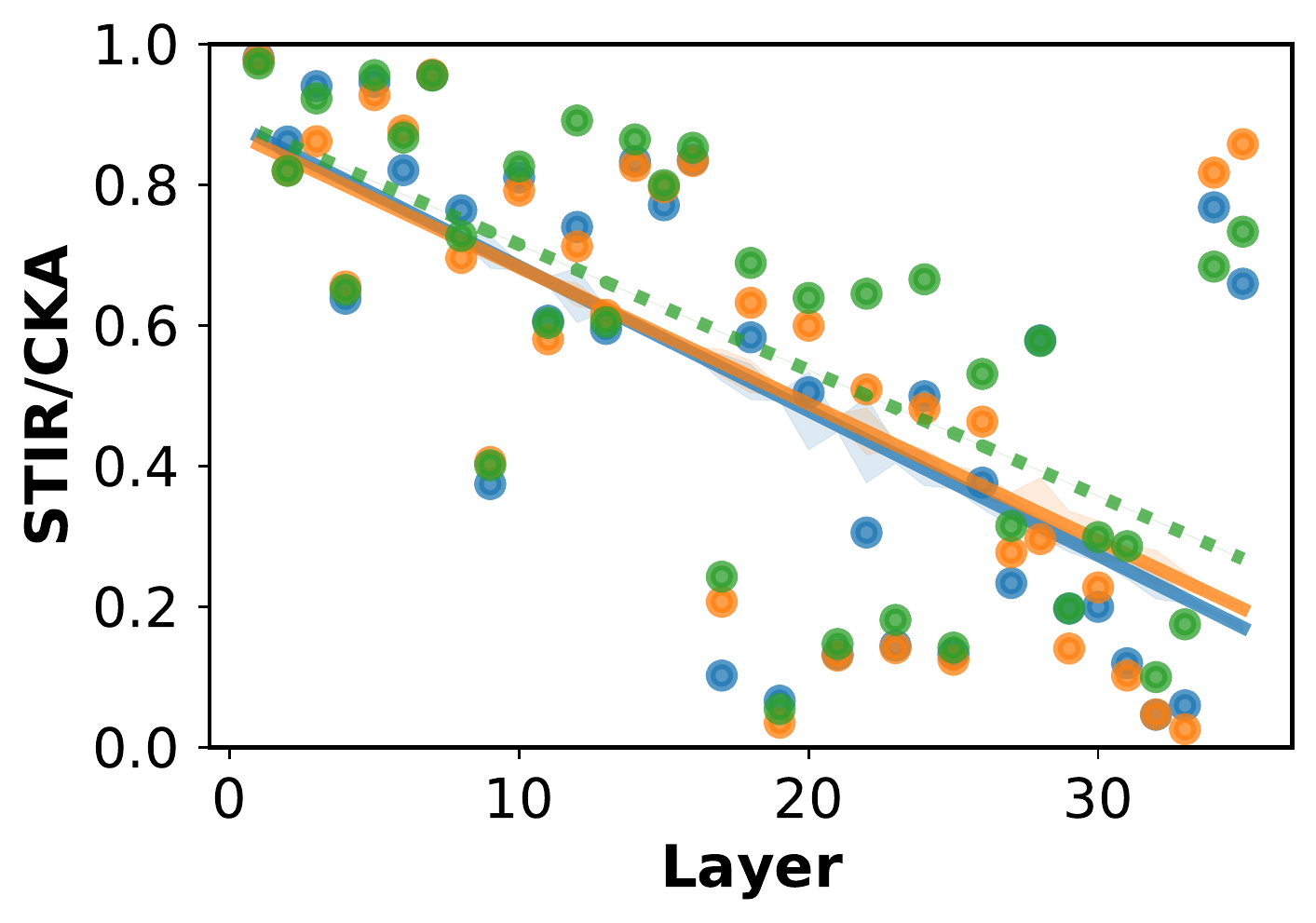}}
    
    \subfloat[\mone = VGG16 on CIFAR10 w ERM\\\mtwo = VGG16 on CIFAR100 w ERM]{\label{fig:diff_datasets_vgg16erm}
        \includegraphics[width=0.35\linewidth]{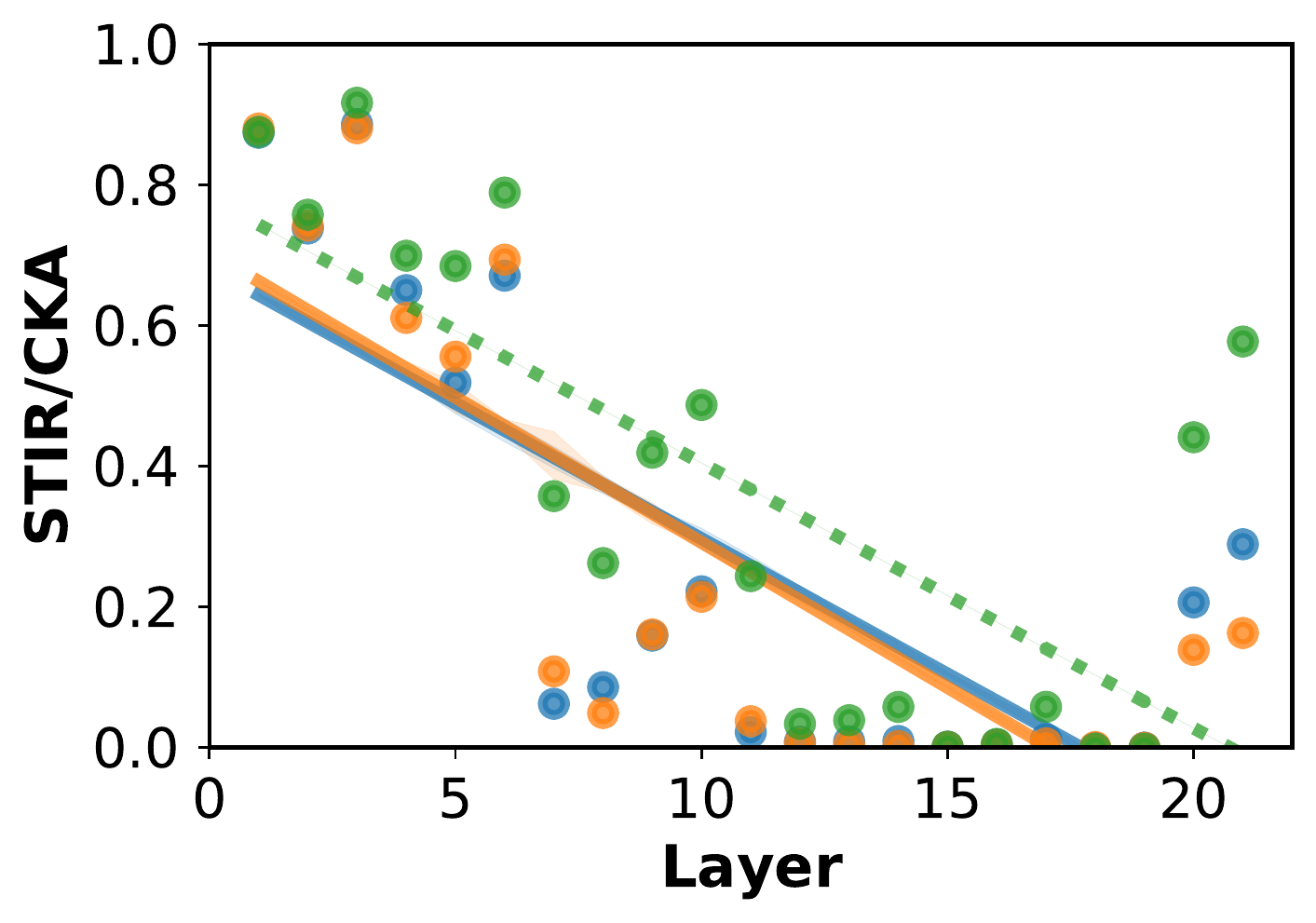}}
    \subfloat[\mone = VGG16 on CIFAR10 w AT\\\mtwo = VGG16 on CIFAR100 w AT]{\label{fig:diff_datasets_vgg16at}
        \includegraphics[width=0.35\linewidth]{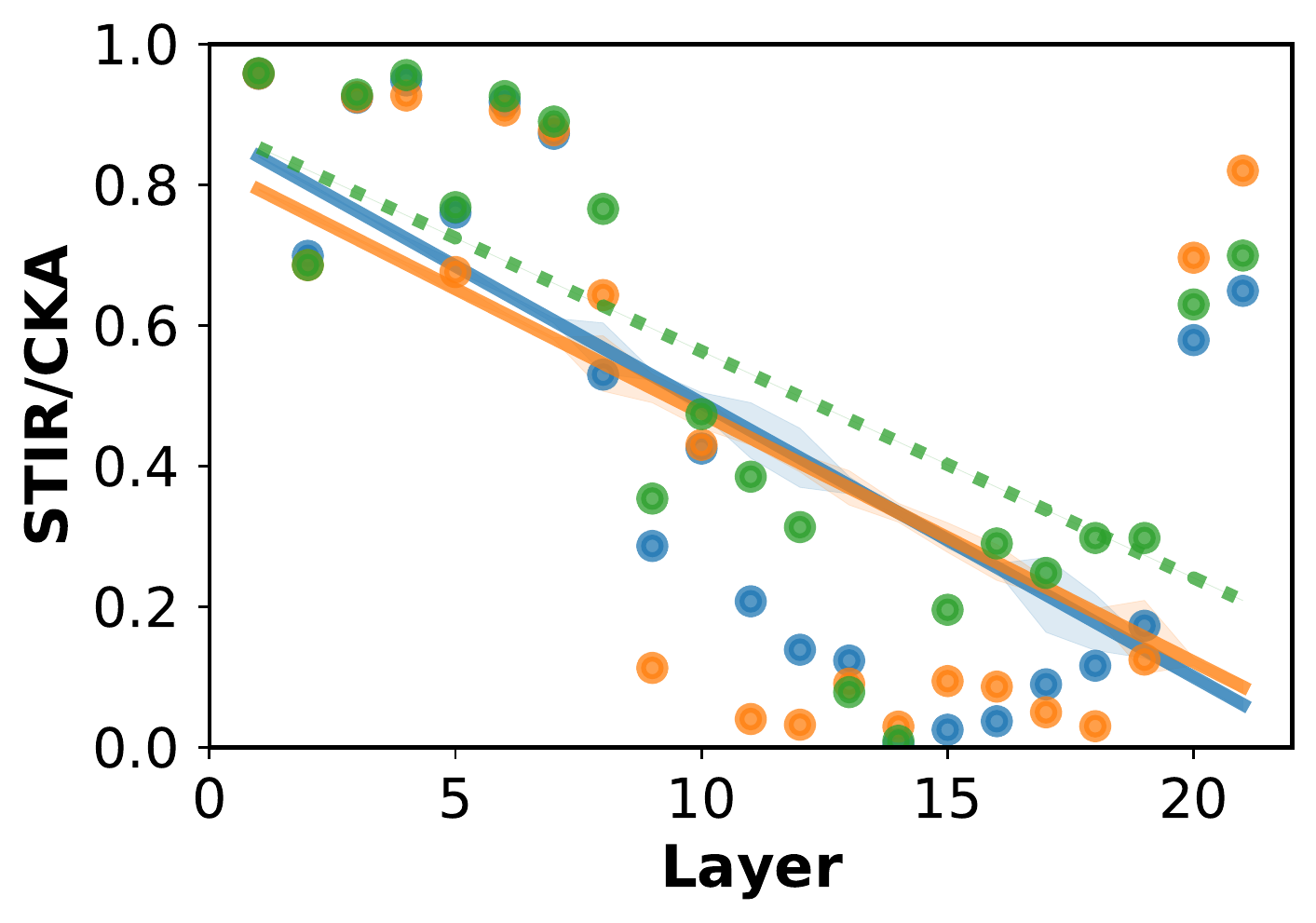}}
    
    \subfloat[\mone = VGG19 on CIFAR10 w ERM\\\mtwo = VGG19 on CIFAR100 w ERM]{\label{fig:diff_datasets_vgg19erm}
        \includegraphics[width=0.35\linewidth]{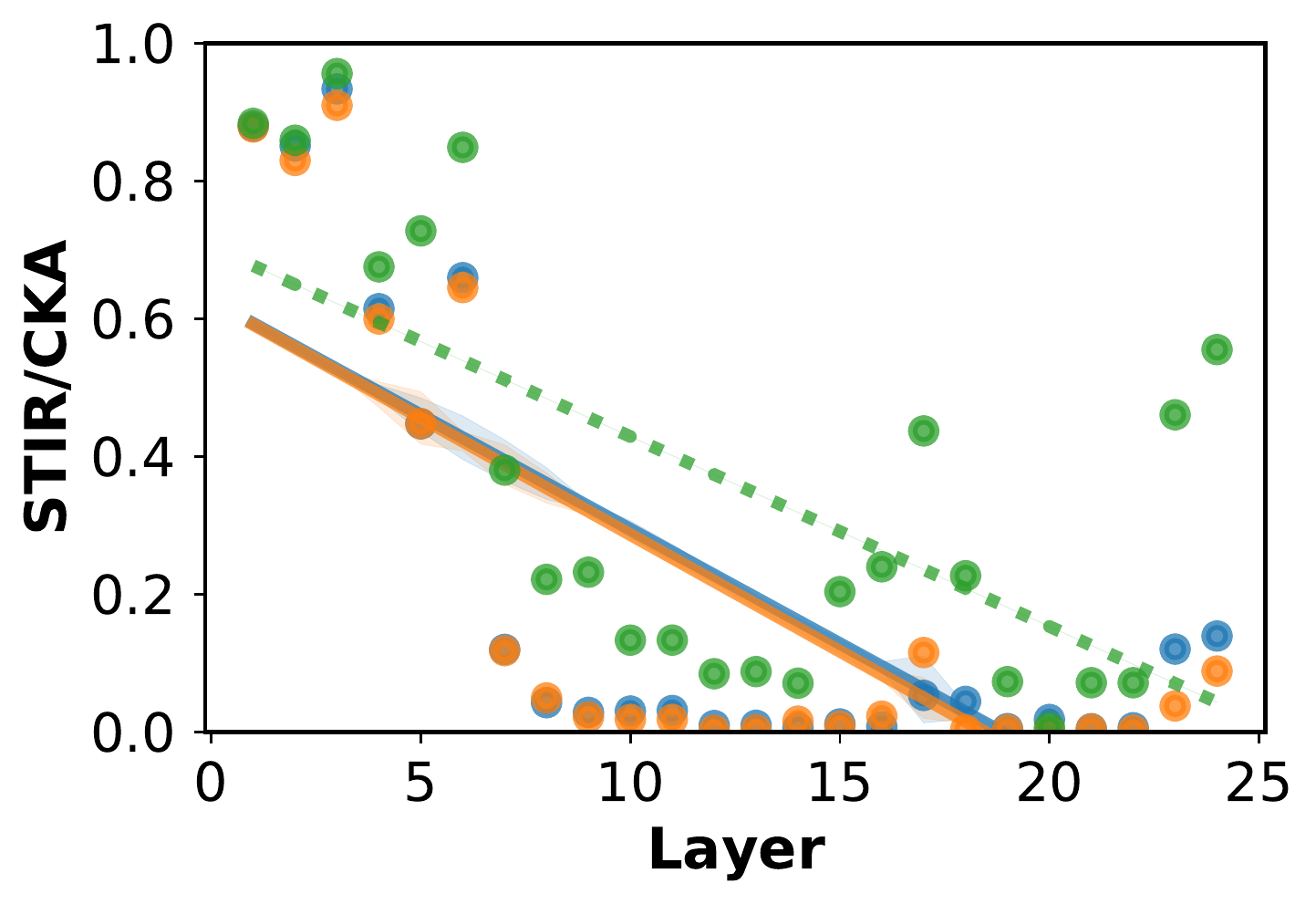}}
    \subfloat[\mone = VGG19 on CIFAR10 w AT\\\mtwo = VGG19 on CIFAR100 w AT]{\label{fig:diff_datasets_vgg19at}
        \includegraphics[width=0.35\linewidth]{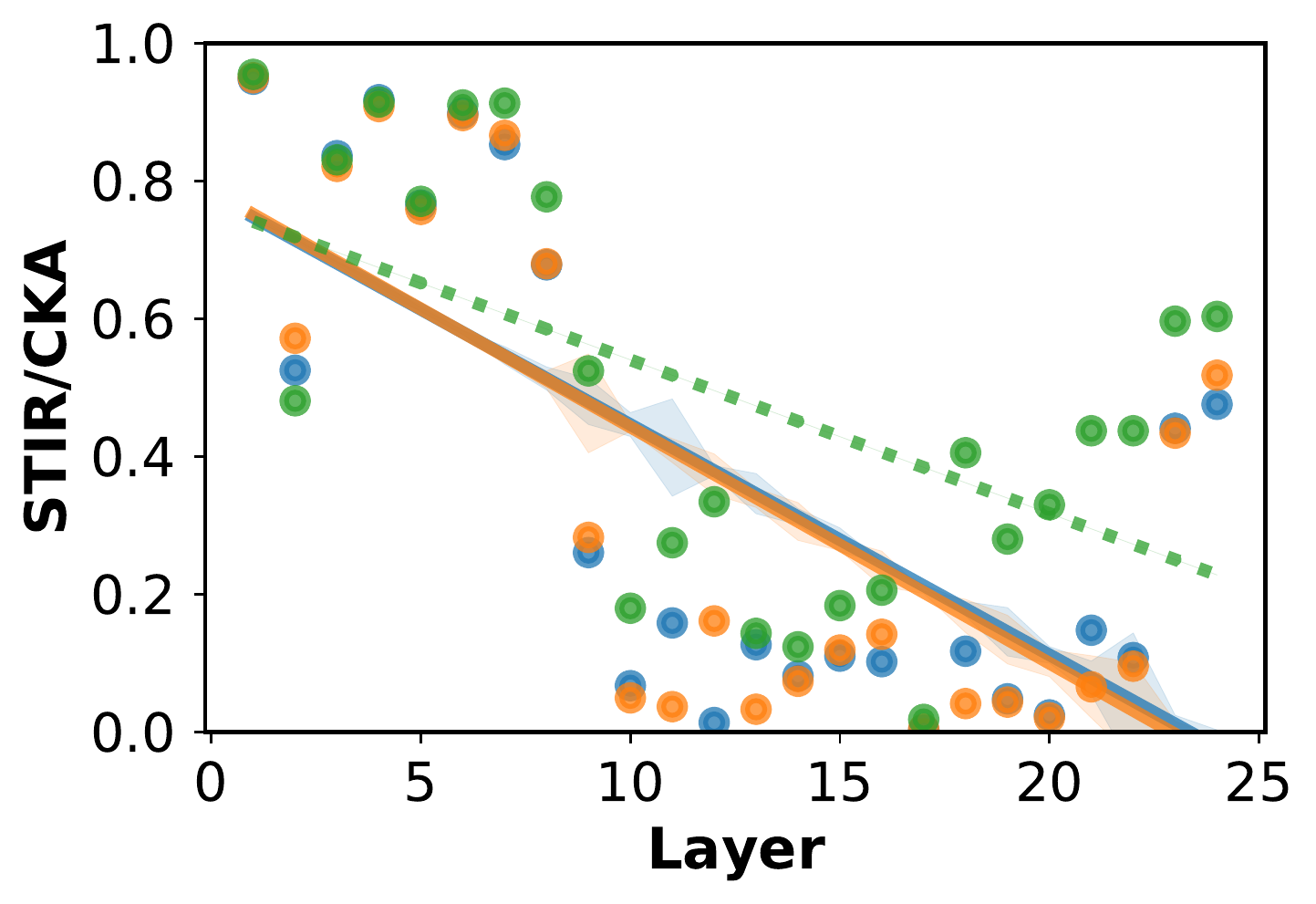}}

\caption{\textbf{[Different Datasets; ResNet34, VGG16 and VGG19 on CIFAR10 and CIFAR100]} Similar to the finding of~\cite{kornblith2019similarity} we find that across datasets, initial layers tend to have more shared invariances. However, we also find that shared invariances increases substantially for all layers with adversarial training (AT).}
\label{fig:diff_datasets_more_models}
\end{figure*}

Fig~\ref{fig:rand_init_more_models}\&\ref{fig:diff_datasets_more_models} shows results for different random initializations and different datasets respectively. We see similar trends hold for deeper models such as ResNet34 and VGG19.

\subsection{Different Architectures, Penultimate Layer}

\begin{figure*}[h!]
    \centering
    
    \subfloat[\stir - completely random seeds]{\includegraphics[width=0.4\linewidth]{figures/sec4_2/STIR_layer_penultimate_small.pdf}}
    \subfloat[\stir - seeds sampled from $\gN(0, 1)$]{\includegraphics[width=0.4\linewidth]{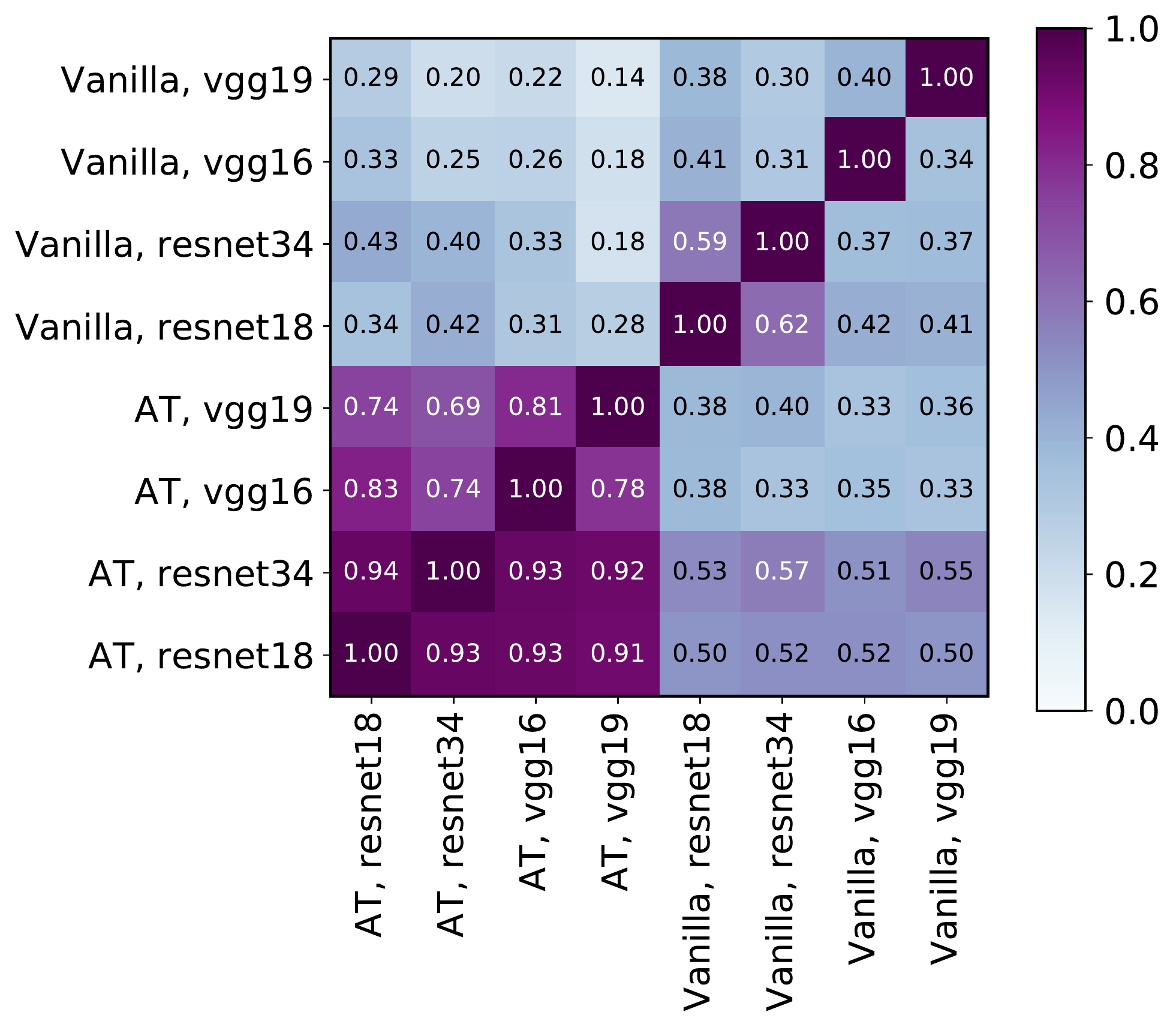}}
    
    \subfloat[\cka]{\includegraphics[width=0.4\linewidth]{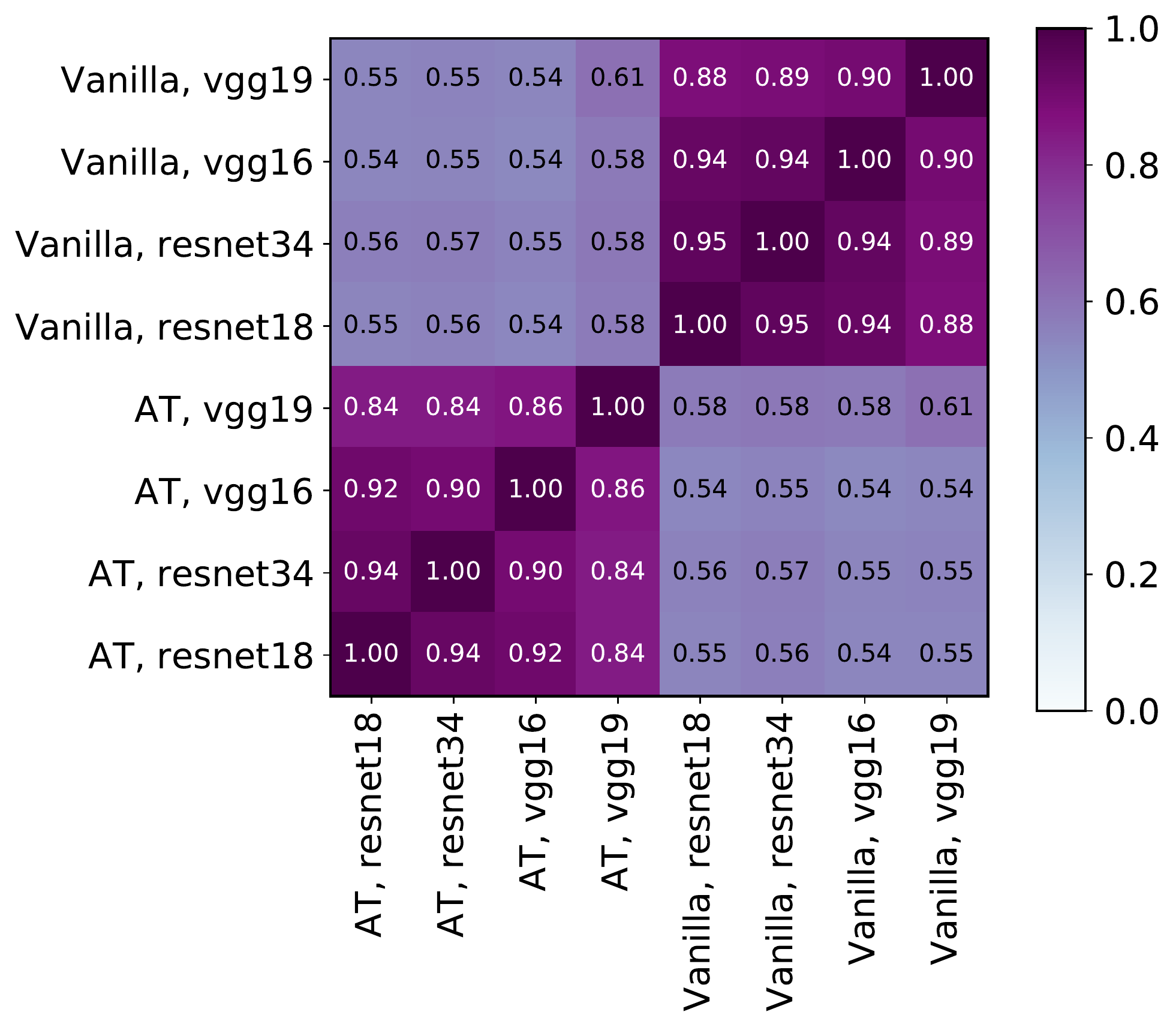}}
    
\caption{\textbf{[Different Architectures, Penultimate Layer Shared Invariances; CIFAR10]} We find that in general ResNets have high shared invariances when trained using the same loss, but this shared invariance drops across training types. We also see that with AT, even models with different architectures converge to high shared invariances. Here additionally we show comparison with \cka which does not provide any faithful estimate of shared robustness and is unformly high for similarly trained models. We also see that this trend holds even when different seeds are used to generate $X'$. }
\label{fig:across_models_stir_more_seeds}
\end{figure*}

Fig~\ref{fig:across_models_stir_more_seeds} shows that trends for \stir hold across different choice of seeds. We also see that in contrast \cka assigns high value across the board and is thus not a faithful measure of shared invariance.

\subsection{Different Adversarial Training Methods}

\begin{figure*}[t!]
    \centering
    
    \subfloat[AT (m$_1$) TRADES (m$_2$)]{\label{fig:diff_adv_vgg16madrytrades}
        \includegraphics[width=0.3\linewidth]{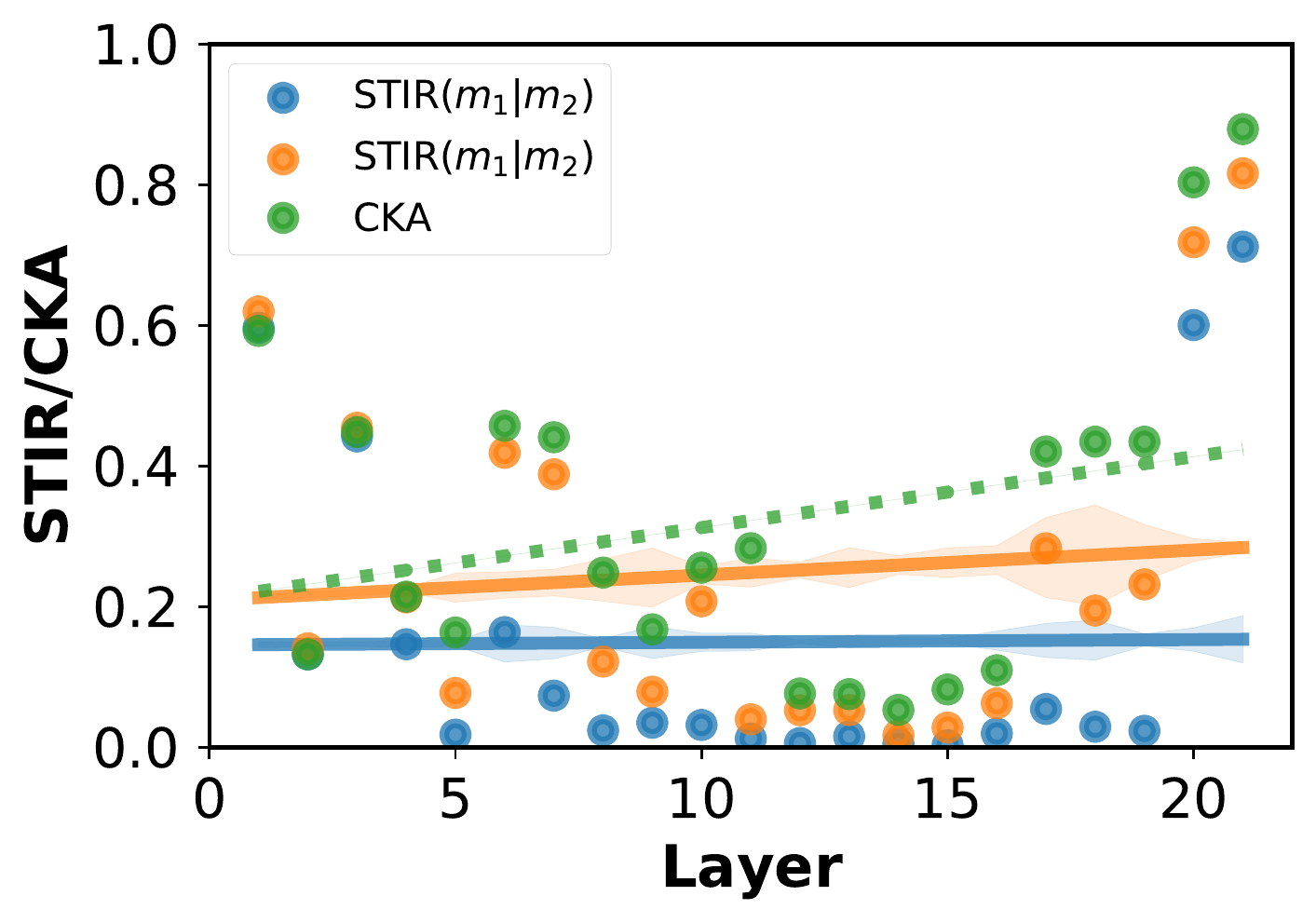}}
    \subfloat[AT (m$_1$) MART (m$_2$)]{\label{fig:diff_adv_vgg16madrymart}
        \includegraphics[width=0.3\linewidth]{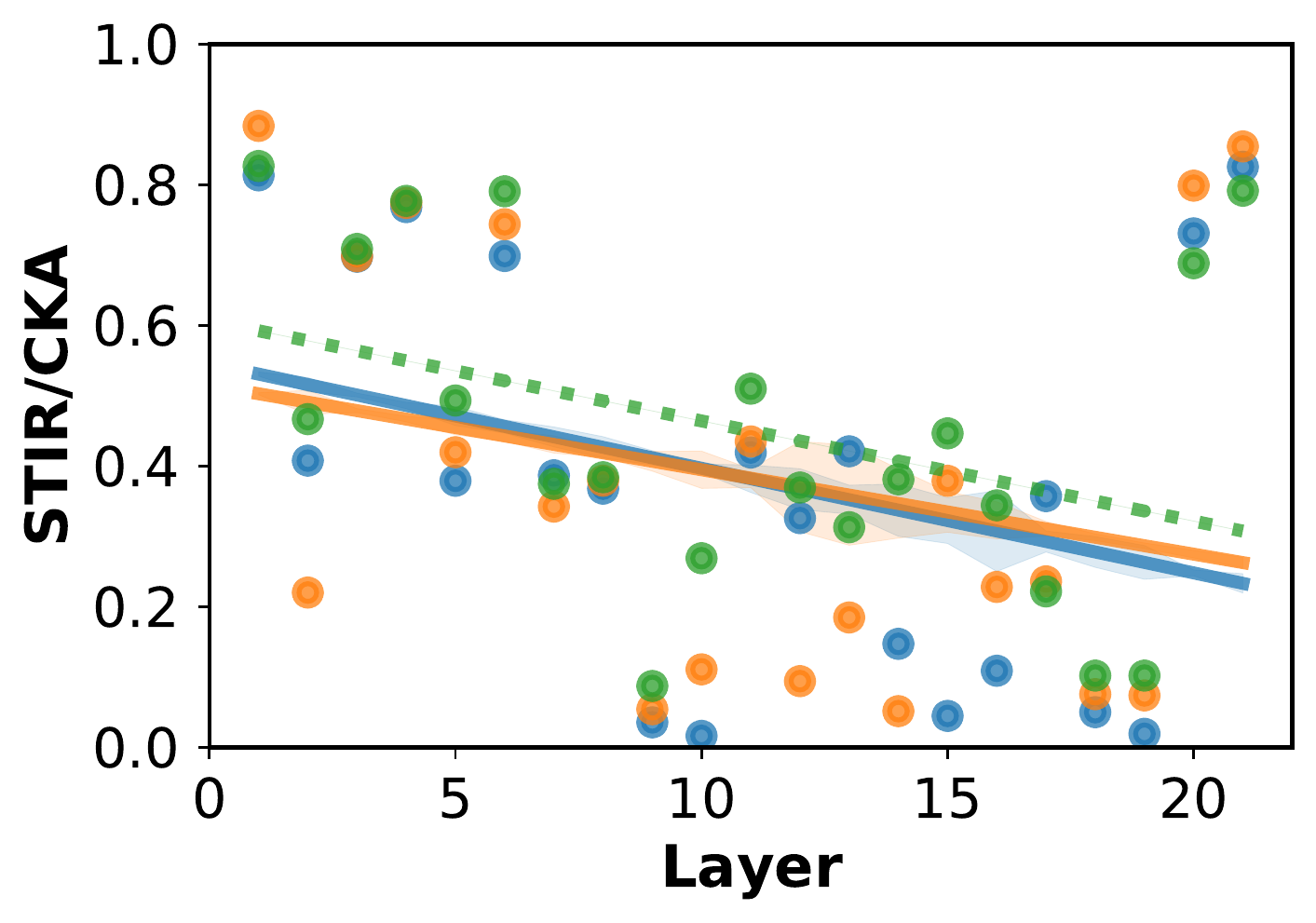}}
    \subfloat[TRADES (m$_1$) vs MART (m$_2$)]{\label{fig:diff_adv_vgg16tradesmart}
        \includegraphics[width=0.3\linewidth]{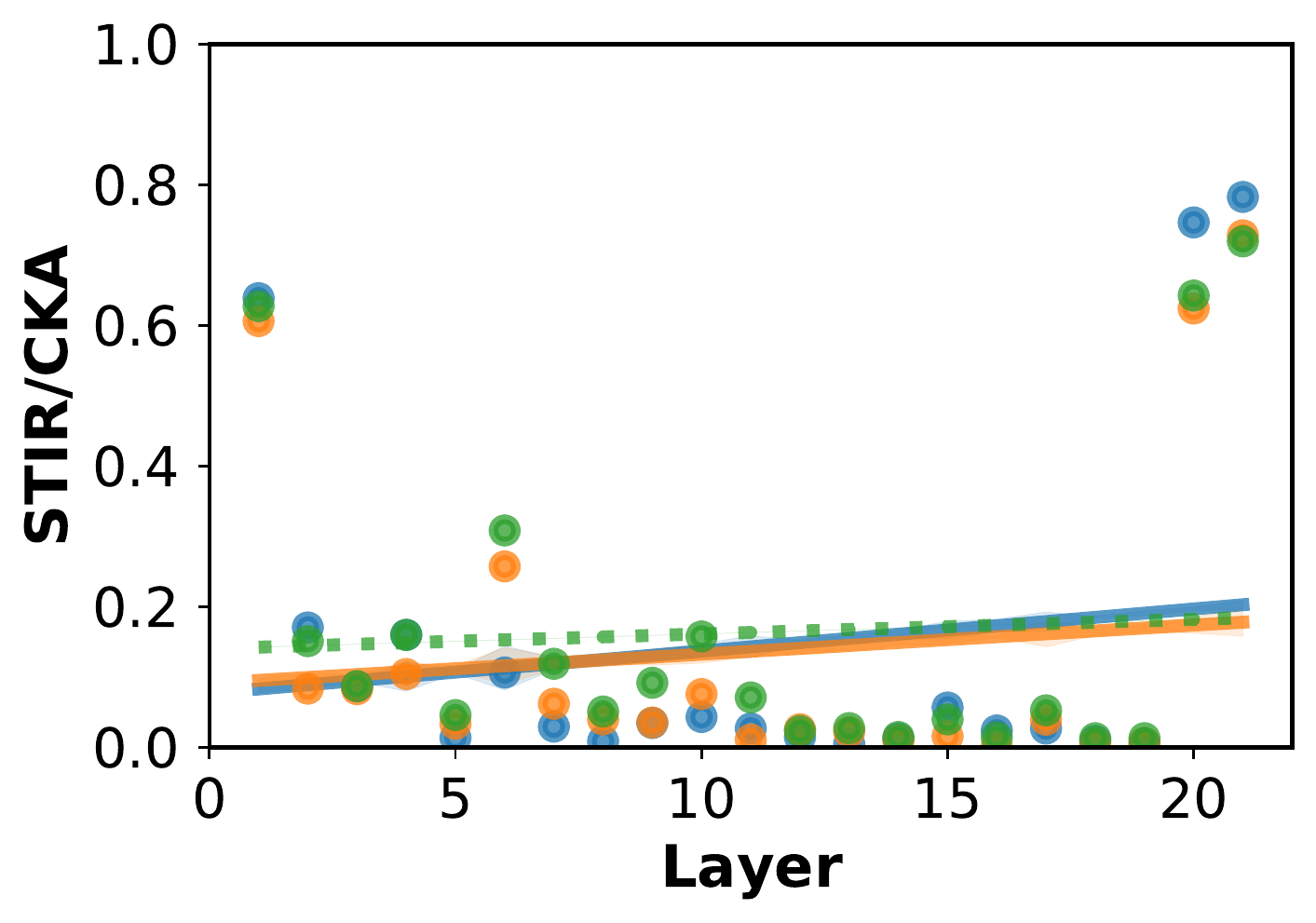}}
\caption{\textbf{[Different Types of Adversarially Robust Training; VGG16 on CIFAR10]} We se much high levels of shared invariance for VGG16 than for ResNet18. Even for VGG16, AT and MART achieve $ell_p$ ball robustness in different ways.}
\label{fig:diff_adv_more_models}
\end{figure*}

Fig~\ref{fig:diff_adv_more_models} shows results on VGG16. These are much higher than values for ResNet18, thus showing that these methods achieve robustness differently across architectures. For AT and MART, we see lower shared invariances even for VGG16.s

\end{document}